\documentclass[lettersize,journal]{IEEEtran}
\usepackage{amsthm}
\theoremstyle{definition}
\newtheorem{definition}{Definition}

\usepackage{amsmath,amsfonts}
\usepackage{algorithmic}
\usepackage{algorithm}
\usepackage{array}
\usepackage[caption=false,font=normalsize,labelfont=sf,textfont=sf]{subfig}
\usepackage{textcomp}
\usepackage{stfloats}
\usepackage{url}
\usepackage{amssymb}
\usepackage{verbatim}
\usepackage{graphicx}
\usepackage{cite}
\usepackage{hyperref}\hypersetup{
  colorlinks=true,
  linkcolor=blue,
  filecolor=blue,      
  urlcolor=blue,
  citecolor=blue,
}
\usepackage{authblk} 
\usepackage{ulem}
\usepackage{tabularx}
\usepackage{multirow}
\begin{document}

\title{Prompt Transfer for Dual-Aspect Cross Domain Cognitive Diagnosis}

\author[1,2]{Fei Liu\thanks{This work was supported in part by grants from the National Science and Technology Major Project (under grant 2021ZD0111802), the National Natural Science Foundation of China (under grants 62406096, 72188101, 62376086), the China Postdoctoral Science Foundation (Grant  No. 2024M760722), and the Fundamental Research Funds for the Central Universities 
 (under grant JZ2024HGQB0093).}
 \thanks{This work has been submitted to the IEEE for possible publication. Copyright may be transferred without notice, after which this version may no longer be accessible.}
 }
\author[1,2]{Yizhong Zhang}
\author[3]{Shuochen Liu}
\author[4]{Shengwei Ji}
\author[1,2]{Kui Yu}
\author[1,2]{Le Wu}

\affil[1]{School of Computer Science and Information Engineering, Hefei University of Technology, Hefei, China}
\affil[2]{Key Laboratory of Knowledge Engineering with Big Data (the Ministry of Education of China), Hefei, China}
\affil[3]{School of Computer Science and Technology, University of Science and Technology of China, Hefei, China}
\affil[4]{School of Big Data And Artificial Intelligence, Hefei University, Hefei, China}


\markboth{Journal of \LaTeX\ Class Files,~Vol.~1, No.~1, November~2024}
{Shell \MakeLowercase{\textit{et al.}}: A Sample Article Using IEEEtran.cls for IEEE Journals}


\maketitle

\newcommand{\shortname}{PromptCD}

\begin{abstract}
Cognitive Diagnosis (CD) aims to evaluate students' cognitive states based on their interaction data, enabling downstream applications such as exercise recommendation and personalized learning guidance. However, existing methods often struggle with accuracy drops in cross-domain cognitive diagnosis (CDCD), a practical yet challenging task. While some efforts have explored exercise-aspect CDCD, such as cross-subject scenarios, they fail to address the broader dual-aspect nature of CDCD, encompassing both student- and exercise-aspect variations. This diversity creates significant challenges in developing a scenario-agnostic framework.
To address these gaps, we propose PromptCD, a simple yet effective framework that leverages soft prompt transfer for cognitive diagnosis. PromptCD is designed to adapt seamlessly across diverse CDCD scenarios, introducing PromptCD-S for student-aspect CDCD and PromptCD-E for exercise-aspect CDCD. Extensive experiments on real-world datasets demonstrate the robustness and effectiveness of PromptCD, consistently achieving superior performance across various CDCD scenarios.
Our work offers a unified and generalizable approach to CDCD, advancing both theoretical and practical understanding in this critical domain. The implementation of our framework is publicly available at \url{https://github.com/Publisher-PromptCD/PromptCD}. 
\end{abstract}

\begin{IEEEkeywords}
Educational Data Mining, Cognitive Diagnosis, Cross-Domain, Prompt Transfer.
\end{IEEEkeywords}

\section{Introduction}
\label{introduction}
\IEEEPARstart{C}{ognitive} diagnosis aims to assess students' proficiency based on their historical interactions \cite{fcs2023new-dev,wang2020ncdm,ma2022kscd}. It is a crucial task in the field of educational data mining, which can support many downstream tasks like exercise recommendation \cite{liu2023meta, huang2019exploring}, learning guidance \cite{wu2023contrastive,fan2022interpretable, bi2020adaptive,chang2015junyi,wang2021nips2020ec}, and computerized adaptive testing \cite{cat_survey}.

In recent efforts, the primary focus has been on enhancing the accuracy of cognitive diagnosis models \cite{wang2024surveymodelscognitivediagnosis}. Specifically, these models aim to learn the characteristics of students and exercises from training data and utilize these learned representations to predict scores on test data \cite{zhou2021ecd, gao2021rcd, wang2022neuralcd, wang2021cdgk, li2022hiercdf, tong2022icd}. Despite advancements in these models, they rely on the assumption that student and exercise characteristics are consistent across training and test data, which can be referred to as \textit{in-domain cognitive diagnosis}.
However, the assumption mentioned above is strict in practice. Students and exercises with diverse characteristics create diverse domains. For instance, students from different schools or countries, while exercises span various subjects. Thus, considering \textit{\textbf{C}ross-\textbf{D}omain \textbf{C}ognitive \textbf{D}iagnosis (CDCD)} for students or exercises with various characteristics is more practical, however,
posing many challenges to existing models. 

\begin{figure}[!t]
  \centering
  \includegraphics[width=1\linewidth]{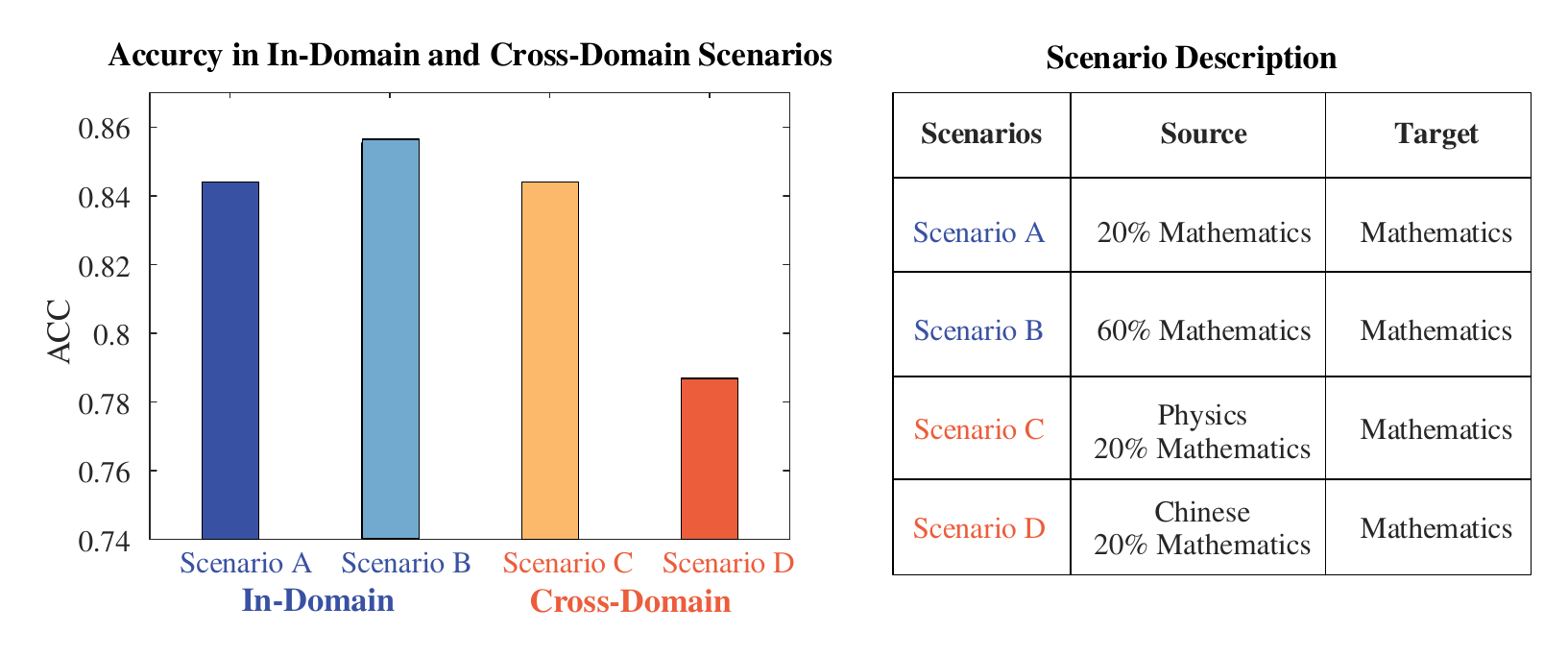}
  \captionsetup{skip=0pt}
  \caption{Performance comparison of MIRT in in-domain scenarios A and B versus cross-domain scenarios C and D.
  }
  \label{intro1}
\end{figure}

\begin{figure}[!t]
  \centering
  \includegraphics[width=1\linewidth]{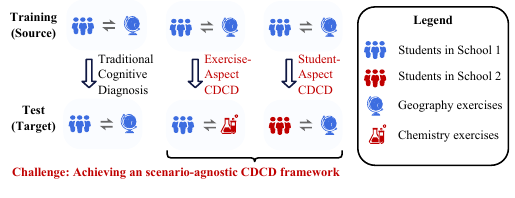}
  \captionsetup{skip=0pt}
  \caption{Traditional cognitive diagnosis versus cross-domain cognitive diagnosis. 
  }
  \label{intro2}
\end{figure}

First, representations of students or exercises learned from training data (source domains) cannot be directly applied to testing data (target domains), leading to a sharp decline in diagnostic accuracy. A straightforward approach to this limitation is to retrain the model. However, retraining solely on new domain data often results in overfitting and catastrophic forgetting, severely compromising the model’s generalization capabilities. Alternatively, retraining with all existing and new data is computationally intensive and impractical for real-world applications. 
To illustrate these limitations, we conducted experiments using MIRT \cite{reckase2009mirt} on the SLP dataset \cite{lu2021slp} in in-domain (A and B) and cross-domain (C and D) scenarios. Figure \ref{intro1} provides detailed scenario descriptions, and the results highlight the following:
1) Cross-Domain Challenges: C and D performed significantly worse than A and B, indicating the struggles of traditional models in CDCD scenarios.
2) Overfitting Risks: A performed worse than B, highlighting the impact of overfitting when retraining with limited data.
3) Sensitivity to Domain Differences: D significantly underperformed C due to greater distributional differences between Chinese and mathematics compared to physics and mathematics, underscoring the sensitivity of traditional models to source domains.
These observations demonstrate that retraining cognitive diagnosis models is impractical, necessitating a better approach to mitigate accuracy declines in CDCD.

Second, CDCD scenarios are inherently complex due to the diverse characteristics of students and exercises, creating a variety of domains. As shown in Figure \ref{intro2}, CDCD can be categorized into two types: 1) \textbf{Student-Aspect CDCD}: Differences arise from varying student demographics, such as urban versus rural populations. 2) \textbf{Exercise-Aspect CDCD}: Variations occur across subject domains, such as mathematics and physics.
Unfortunately, CDCD remains an underexplored area. Existing studies \cite{gao2023leveraging, gao2023zero} have primarily focused on specific CDCD aspects, proposing scenario-specific models with limited compatibility. Developing a generalizable framework capable of addressing both student- and exercise-aspect CDCD scenarios remains a significant challenge.

In this paper, we propose PromptCD, a simple yet generalizable framework designed to address the challenges of dual-aspect cross-domain cognitive diagnosis (CDCD). Dual-aspect CDCD introduces unique complexities, as knowledge transfer must consider both student-aspect and exercise-aspect scenarios. These scenarios involve distinct challenges: 1) \textbf{Dual-Aspect Entity Diversity}: Students and exercises across domains can be either overlapping or non-overlapping. Overlapping entities require personalized adaptation to maintain consistency, while non-overlapping entities necessitate a generalized representation to ensure effective knowledge transfer.
2) \textbf{Adaptation Across Domains}: Diverse target domains vary significantly in their characteristics, making it difficult to adapt representations while preserving diagnostic accuracy and avoiding issues like overfitting or catastrophic forgetting.
To address these challenges, PromptCD introduces a unified framework leveraging soft prompt transfer, a proven technique in cross-domain tasks across various fields \cite{liu2023pre, ge2023domain, liu2022prompt, wu2022adversarial, zhao2023spc, zhao2022adpl}. Specifically, we design personalized prompts for overlapping entities and shared domain-adaptive prompts for non-overlapping entities. These prompts enhance representation learning and transfer, ensuring robustness in diverse CDCD scenarios. Additionally, PromptCD adopts a two-stage training strategy—pre-training on source domains and fine-tuning on target domains—for efficient and scalable adaptation.
To demonstrate its versatility, we develop PromptCD-S and PromptCD-E, tailored to student-aspect and exercise-aspect CDCD scenarios, respectively.
We summarize the contributions of this paper as follows:
\begin{itemize}
    \item We propose the PromptCD framework, introducing soft prompt transfer and a two-stage training strategy to address dual-aspect CDCD challenges.
    \item We develop PromptCD-S and PromptCD-E, showcasing the framework’s ability to generalize across student- and exercise-aspect scenarios.
    \item Extensive experiments on real-world datasets validate the effectiveness of PromptCD, achieving significant performance improvements over baselines.
\end{itemize}

\section{Preliminaries}


\subsection{Cognitive Diagnosis}
\label{sec_Preliminaries_cd}

Cognitive diagnosis aims to evaluate students' proficiency in knowledge concepts based on their response records $\boldsymbol{L}$. This task involves modeling the interaction between student features $\boldsymbol{\alpha}$ and exercise features $\boldsymbol{\beta}$ to predict scores. Since students' proficiency is not directly observable, the model is trained to optimize predictive accuracy using the cross-entropy loss $\mathcal{L}_{CE}$. Below, we outline key interaction functions used in classic cognitive diagnosis models:

IRT \cite{embretson2013irt} and MIRT \cite{reckase2009mirt} use the logistic function in a unidimensional and multidimensional manner, respectively. The detailed interaction functions are as follows: $y_{uv}=\frac{1}{1+{e}^{-C*\boldsymbol{D}_v(\boldsymbol{\alpha}_u-\boldsymbol{\beta}_v)}}$ and
 $y_{uv}=\frac{1}{1+{e}^{-\boldsymbol{\alpha}_u^{T}\boldsymbol{\beta}_v+{D}_v}}$,
where ${D}_{v}$ is discrimination of exercise ${v}$. $C$ is a constant. 

NeuralCD \cite{wang2020ncdm,wang2022neuralcd} utilizes neural networks to model the complex interactions between representations of students and exercises as follows: $\boldsymbol{x}_{uv}=\boldsymbol{Q}_{v} \circ (\boldsymbol{\alpha}_u-\boldsymbol{\beta}_v)*{D}_v,y_{uv} = f_{1}(f_{2}(f_{3}(\boldsymbol{x}_{uv})))$,
where $\boldsymbol{Q}_{v} \in \{0,1\}^{1*K}$  indicates whether an exercise is associated with a knowledge concept.
$f_{1},f_{2},f_{3}$ are the fully connected layers with positive weights to ensure monotonicity.

KSCD \cite{ma2022kscd} further explores the impact of potential associations between knowledge concepts on diagnostic results,  shown as follows: $\boldsymbol{\alpha}_{uc}' = \phi \left( f_{sk} (\boldsymbol{\alpha}_u \oplus \mathbf{h}_c^K) \right), \boldsymbol{\beta}_{vc}' = \phi \left( f_{ek} (\boldsymbol{\beta}_v \oplus \mathbf{h}_c^K) \right),y_{uv} = \phi \left( \frac{1}{n_v} \sum_{c=1}^C \mathbf{Q}_{vc} \times f{se} (\boldsymbol{\alpha}_{uc}' - \boldsymbol{\beta}_{vc}') \right),$
where $\mathbf{h}_c^K$ represents the initialized embedding representations of knowledge concept $c$. $\mathbf{Q}_{vc}$ is the knowledge relevance vector $\mathbf{Q}_v$ of the concept $c$. ${n_v}$ indicates the number of knowledge concepts contained in exercise ${e_v}$. ${\phi}$ is the activation function. ${f_{se}, f_{sk}, f_{ek}}$ are linear transformation functions that correspond to different fully connected layers.

\subsection{Cross-Domain Cognitive Diagnosis (CDCD)}
\label{sec:CDCD}
Consider $|\boldsymbol{S}|$ source domains $\{\boldsymbol{S}_{1}, \boldsymbol{S}_{2}, \ldots, \boldsymbol{S}_{|\boldsymbol{S}|}\}$ and $|\boldsymbol{T}|$ target domains $\{\boldsymbol{T}_{1}, \boldsymbol{T}_{2}, \ldots, \boldsymbol{T}_{|\boldsymbol{T}|}\}$. Let $\boldsymbol{LS}_{s}$ and $\boldsymbol{LT}_{t}$ denote the interaction records for source domain $\boldsymbol{S}_{s}$ and target domain $\boldsymbol{T}_{t}$, respectively, where $s \in \{1, 2, \ldots, |\boldsymbol{S}|\}$ and $t \in \{1, 2, \ldots, |\boldsymbol{T}|\}$.
The CDCD task aims to identify and leverage cognitive patterns and learning structures that generalize across domains, enhancing the model's performance in the new domain $\boldsymbol{T}_{t}$. By leveraging the abundant data $\boldsymbol{LS}_{1}, \boldsymbol{LS}_{2}, \ldots,\boldsymbol{LS}_{|S|}$ from source domains, CDCD can rapidly establish cognitive diagnosis models in target domain $\boldsymbol{T}_{t}$ with few-shot data $\boldsymbol{LT}_{t}^{few} \subset \boldsymbol{LT}_{t}$. 

\begin{figure}[!t]
  \centering
  \includegraphics[width=0.95\linewidth]{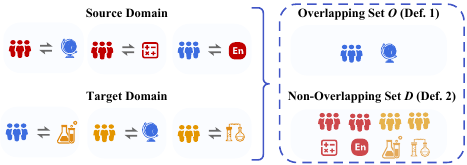}
    \captionsetup{skip=1pt}
  \caption{
Illustration of overlapping and non-overlapping sets in CDCD.
}
  \label{fig_overlap}
\end{figure}

To facilitate the introduction of the subsequent framework, we define sets $\boldsymbol{O}$ and $\boldsymbol{D}$ to represent the overlapping and non-overlapping entities between the source and target domains, as depicted in Figure \ref{fig_overlap}. The entity here refers to students or exercises.
For instance, in the exercise-aspect CDCD scenarios, there exists non-overlapping groups of exercises, defined as $\boldsymbol{D}$. 
Conversely, we define the set of overlapping students as $\boldsymbol{O}$, which allows the transfer of cross-domain information. 



\begin{definition}[Overlapping Set]
The overlapping set $\boldsymbol{O}$ represents the entities that exist in both the source and target domains, which is defined as:
\begin{equation}
\boldsymbol{O} = \bigcup_{s=1}^{|\boldsymbol{S}|} \boldsymbol{S}_s \cap \bigcup_{t=1}^{|\boldsymbol{T}|} \boldsymbol{T}_t
\end{equation}
\end{definition}
\begin{definition}[Non-Overlapping Set]
Let $\boldsymbol{\Omega}$ be the universal set of all entities in both the source and target domains. The non-overlapping set $\boldsymbol{D}$ is defined as the complement of $\boldsymbol{O}$ with respect to the universal set $\boldsymbol{\Omega}$:
\begin{equation}
\boldsymbol{D} = 
\boldsymbol{\Omega} \setminus \boldsymbol{O} = 
\left(\bigcup_{s=1}^{|\boldsymbol{S}|} \boldsymbol{S}_s \cup \bigcup_{t=1}^{|\boldsymbol{T}|} \boldsymbol{T}_t\right) \setminus \boldsymbol{O}
\end{equation}
\end{definition}

\begin{figure*}[!t]
  \centering
  \includegraphics[width=1\linewidth]{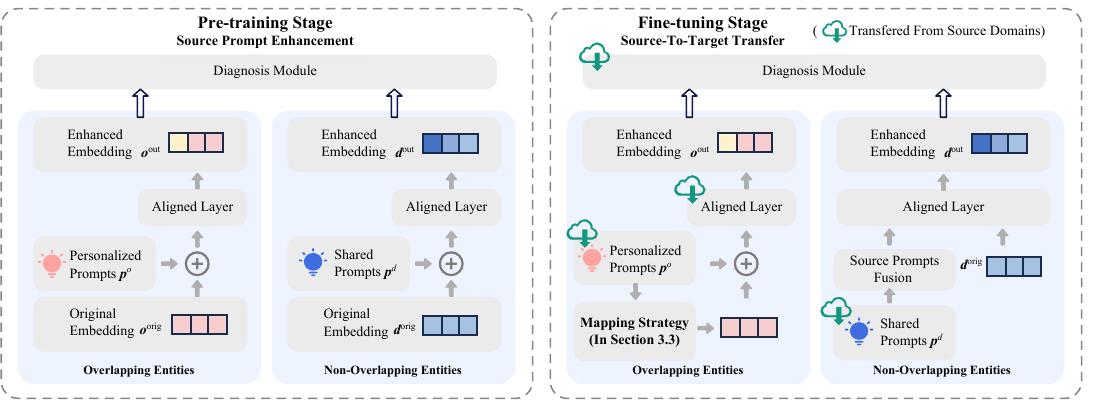}
    \captionsetup{skip=3pt}
  \caption{
Overall architecture of the proposed PromptCD framework, including the pre-training and fine-tuning stages.
}
  \label{fig_framework}
\end{figure*}
\section{Proposed Framework}
\label{sec_framework}
In this section, we propose the scenario-agnostic~\shortname~, applicable to both student- and exercise-aspect CDCD scenarios.

\subsection{Overall Architecture}
\label{sec_Architecture}
The overall architecture is shown in Figure \ref{fig_framework}. Our two-stage framework abstracts scenario-agnostic features and unified learning strategies, enabling rapid adaptation to new domains.
The pseudo-code for ~\shortname~ is presented in Algorithm \ref{alg_pre_training}.

\textbf{Pre-training Stage.} In Section \ref{sec:Source Prompt}, we present an exposition of the personalized and shared prompts, as well as the processing strategies for entity representations. During the pre-training stage, the prompts are updated using the data $\boldsymbol{LS}_{s}$ ($s \in \{1, 2, \dots, |S|\}$) from the source domains.

\begin{algorithm}[!t]
\small
  \renewcommand{\algorithmicrequire}{\textbf{Input:}}
\renewcommand\algorithmicensure {\textbf{Output:}}
  \caption{\shortname}
  \label{alg_pre_training}
  \begin{algorithmic}[1]
    \STATE \textbf{Input:}  cognitive diagnosis model $\mathcal{M}$, records $\boldsymbol{LS}$ for pre-training and $\boldsymbol{LT}_{t}^{few}$ in target doamin $t$ for fine-tuning.
    \STATE \textbf{Output:} fine-tuned model $\mathcal{M}$, the transfer prompts $\boldsymbol{\hat p}^{o}$ and $\boldsymbol{\hat p}^{d}_{t}$.
    \STATE ---\textbf{Pre-training Stage}---
    \WHILE{$e_{1} \leqslant Epoch_{Pretrain}$}
    \FOR{$\boldsymbol{LS}_{s} \in \{\boldsymbol{LS}_{1}, \boldsymbol{LS}_{2},...,\boldsymbol{LS}_{|\boldsymbol{S}|}$\}}
    \STATE Initialize embeddings $\boldsymbol{o}_{s}^{\text{orig}}$, $\boldsymbol{d}_{s}^{\text{orig}}$,  prompts $\boldsymbol{p}^{o}$, $\boldsymbol{p}^{d}$ and $\mathcal{M}$;
    \STATE Enhance the representation of entities in Eq.\eqref{eq_source_con}  and Eq.\eqref{eq_source_fc};
    \STATE Input $\boldsymbol{o}_{s}^{\text{out}}$ and $\boldsymbol{d}_{s}^{\text{out}}$ to $\mathcal{M}$ to predict scores $\boldsymbol{y}_{s}$;
    \STATE Calculate the loss using $\boldsymbol{LS}_{s}$ to update the model;
    \ENDFOR
    \ENDWHILE
    \STATE ---\textbf{Fine-Tuning Stage}---
    \WHILE{$e_{2} \leqslant Epoch_{Finetune}$}
    \STATE Initialize the entities $\boldsymbol{o}_{t}^{\text{orig}}$, $\boldsymbol{d}_{t}^{\text{orig}}$;
    \STATE Obtain the transfer prompts $\boldsymbol{\hat p}^{o}$ and $\boldsymbol{\hat p}^{d}_{t}$ in Eq.\eqref{eq_target_O_transfer} and Eq.\eqref{eq_target_D_transfer});
    \STATE Activate improvement policy in Eq.\eqref{eq_target_O_transform};
    \STATE Enhance the representations in a manner similar to pre-training;
    \STATE Input $\boldsymbol{o}_{t}^{\text{out}}$ and $\boldsymbol{d}_{t}^{\text{out}}$ to $\mathcal{M}$ to predict scores $\boldsymbol{y}_{t}$;
    \STATE Calculate the loss using $\boldsymbol{LT}_{t}^{few}$ to update the model.
  \ENDWHILE
  \end{algorithmic}
\end{algorithm}

\textbf{Fine-tuning Stage.} In Section \ref{sec:Source-To-Targe}, we outline the prompt transfer process and introduce a variant that enhances adaptation. After pre-training, we fine-tune the trainable parameters using few-shot data $\boldsymbol{LS}_{t}^{few}$ from the target domain $\boldsymbol{T}_{t}$, thereby transferring knowledge from the source domains and adapting to the target domain's distribution.


\subsection{Source Prompt Enhancement}
\label{sec:Source Prompt}

Considering the characteristics of overlapping and non-overlapping entities in cross-domain scenarios, relying solely on entity representations may not accommodate the diverse interactions across different domains. Consequently, we designed two types of learnable soft prompts—personalized prompts and shared prompts—to establish connection between the source and target domains by associating these prompts with different entity representations.


Specifically,  personalized prompts are tailored for individual entity within the overlapping set $\boldsymbol{O}$. Since the overlapping entities are identical in both the source and target domains, each entity can be associated with a personalized prompt, allowing for the transfer of more information. In contrast, shared prompts are utilized by all entities in each domain within the non-overlapping set $\boldsymbol{D}$, ensuring a common representation for domain-specific knowledge. The resulting composite representations can be expressed as:
\begin{equation}
\begin{split}
\label{eq_source_con}
   \boldsymbol{o}_{k, i}^{\text{cat}} = [\boldsymbol{p}_{i}^{o}, \boldsymbol{o}_{k, i}^{\text{orig}}],\quad
   \boldsymbol{d}_{k, j}^{\text{cat}} = [\boldsymbol{p}_{k}^{d}, \boldsymbol{d}_{k, j}^{\text{orig}}],
\end{split}
\end{equation}
where $\boldsymbol{p}_{i}^{o}$ is the personalized prompt for each individual overlapping entity ${i}$, and $\boldsymbol{p}_{k}^{d}$ is the shared prompt for domain-specific entities. $\boldsymbol{o}_{k, i}^{\text{orig}}$ and $\boldsymbol{d}_{k, j}^{\text{orig}}$ respectively denote original embedding of single entity in source domain $\boldsymbol{S}_{k}(k \in 1,2,...,|\boldsymbol{S}|)$ through random initialization, while $\boldsymbol{o}_{k, i}^{\text{cat}}$ and $\boldsymbol{d}_{k, i}^{\text{cat}}$ denote their corresponding representations after concatenation with prompts.

To integrate the concatenated features and extract the joint information, we utilize a fully connected layer defined as the operator $\text{Linear}$, which has different trainable parameters depending on the types of entities:
\begin{equation}
\begin{split}
\label{eq_source_fc}
\boldsymbol{o}_{k}^{\text{out}} = \text{Linear}_{o}(\boldsymbol{o}_{k}^{\text{cat}}), \quad
\boldsymbol{d}_{k}^{\text{out}} = \text{Linear}_{d}(\boldsymbol{d}_{k}^{\text{cat}}).
\end{split}
\end{equation}
$\boldsymbol{o}_{k}^{\text{out}}$ and $\boldsymbol{d}_{k}^{\text{out}}$ denote final source-domain representations, aligned with the original embeddings. The processed representations are then input into the cognitive diagnosis model to predict scores.

\subsection{Source-To-Target Transfer}
\label{sec:Source-To-Targe}

In the fine-tuning stage, learned prompts are adapted to the target domains. 
Personalized prompts, designed for overlapping entities across domains, are transferred on a one-to-one basis in Eq.(\ref{eq_target_O_transfer}) to maintain the integrity of cross-domain connection information, as these prompts are also relevant to the target domains.
 \begin{equation}
\begin{split}
\label{eq_target_O_transfer}
   & \boldsymbol{\hat p}^{o}_{i}=\boldsymbol{p}^{o}_{i},
\end{split}
\end{equation}
To effectively capture the commonalities across different source domains, the shared prompts should be concatenated and mapped back to their original dimensions:
\begin{equation}
\begin{split}
\label{eq_target_D_transfer}
   & \boldsymbol{\hat p}^{d}_{t}=\text{Linear}_{s2t} (\boldsymbol{p}^{d}_{1}\oplus \boldsymbol{p}^{d}_{2}\oplus ...\oplus \boldsymbol{p}^{d}_{|\boldsymbol{S}|})
\end{split}
\end{equation}
where $\boldsymbol{\hat p}^{d}_{t}$ denotes the shared prompts in target domain $\boldsymbol{T}_{t}$. The operation $\oplus$ indicates the concatenation of prompts from source domains, which are subsequently transformed by the linear mapping function $\text{Linear}_{s2t}$.

The final representations $\boldsymbol{o}_{t}^{\text{out}}$ and $\boldsymbol{d}_{t}^{\text{out}}$ in target domain $\boldsymbol{T}_{t}$ are obtained by processing the original embeddings $\boldsymbol{o}_{t}^{\text{orig}}$ and $\boldsymbol{d}_{t}^{\text{orig}}$ through operations analogous to those described in Eq.\eqref{eq_source_con} and Eq.\eqref{eq_source_fc}, involving interactions with the transferred prompts. 

\subsubsection*{Prompt-to-Representation Mapping}
\label{sec_improvement}
We propose a variant that leverages the cross-domain information characteristics of $\boldsymbol{\hat p}^{o}$, which are learned from the overlapping set $\boldsymbol{O}$. This strategy employs a linear layer to learn the mapping relationship between personalized prompts and shallow representations of entities in the target domains:
\begin{equation}
\begin{split}
\label{eq_target_O_transform}
   & \boldsymbol{o}^{\text{orig}}_{t}=\text{Linear}_{init}(\boldsymbol{\hat p}^{o})
\end{split}
\end{equation}
where $\boldsymbol{{o}^{\text{orig}}_{t}}$ is original representations from $\boldsymbol{O}$ in target domain $\boldsymbol{T}_{t}$. $\text{Linear}_{init}$ comprises trainable parameters that are optimized using the interaction data in the target domains. This strategy supersedes random initialization, leveraging available data to access potential original information.




\section{Proposed Instantiations}
\label{sec:instantiations}
Based on the unified framework above, we instantiate specific scenarios to illustrate its application in the following two scenarios. The pseudo-codes for the proposed PromptCD-S and PromptCD-E instantiations are detailed Algorithms \ref{alg_promptCDS} and \ref{alg_promptCDE} in the Supplements.

\subsection{Student-Aspect CDCD: PromptCD-S}
Student-aspect CDCD focuses on a cross-school scenario where $\boldsymbol{O}$ and $\boldsymbol{D}$ respectively represent a set of exercises and students. This indicates that the source and target domains have overlapping students. Each exercise item has a specific personalized prompt $\boldsymbol{p}^{o}_{exer}$.
We use personalized prompts to uncover the basic requirements of exercise, which are the same for students from different schools. 

In this scenario, each school has a corresponding shared prompt $\boldsymbol{p}^{d}_{sch}$, which reflects the collective performance of students from the common school. Then we employ the PromptCD to concatenate representations of students and exercises with their prompts in Eq.\eqref{eq_source_con} and input them into the cognitive diagnosis model for interaction. By optimizing the cross-entropy loss derived from predicted scores and ground-truth from the source domain response records, PromptCD updates the prompts to extract information across source domains.

To accomplish the source-to-target prompt transfer, we concatenate $\boldsymbol{p}^{d}_{sch}$ for different schools in the source domains and map them to the original dimension in Eq.\eqref{eq_target_D_transfer} to capture commonalities. Finally, a few records from the target domains are used to fine-tune $\boldsymbol{\hat p}^{d}_{sch}$ and personalized prompts $\boldsymbol{\hat p}^{o}_{exer}$ trained from source domains, enhancing their accuracy for future predictions.


\subsection{Exercise-Aspect CDCD: PromptCD-E}
Exercise-aspect CDCD, on the other hand, addresses the cross-subject scenario where $\boldsymbol{O}$ and $\boldsymbol{D}$ respectively represent a set of students and exercises. Each student is associated with a personalized prompt $\boldsymbol{p}^{o}_{stu}$ to capture their basic capability across various subjects, such as mathematics or physics.  

Similarly, each subject has a shared prompt $\boldsymbol{p}^{d}_{sub}$ for all exercise items to enhance the understanding of subject-specific knowledge. Employing the pre-training and fine-tuning methodology analogous to PromptCD-S, we derive the ultimate representations of prompts. 

In different scenarios, the meaning and dimensions of entity representations often differ. PromptCD mitigates the sensitivity of existing studies \cite{gao2023leveraging,gao2023zero} to cross-domain data by employing specific prompts to transfer information across domains, thereby aiding the cognitive diagnosis model in predicting scores accurately.

\section{Experiments} \label{Experiments}
To validate the effectiveness of the~\shortname~in cross-domain scenarios, we conducted extensive experiments on real-world datasets, to address the following questions:
\begin{itemize}
    \item \textbf{RQ1}: How does~\shortname~perform in student- and exercise-aspect CDCD scenarios?
    \item \textbf{RQ2}: How efficient are the key components in~\shortname?
    \item \textbf{RQ3}: Can feature visualization demonstrate the effectiveness of prompts in enhancing cross-domain representations?
    \item \textbf{RQ4}: How to conduct personalized learning guidance using~\shortname?
\end{itemize}


\subsection{Experimental Settings}
We present the experimental setup, including the datasets, baselines, metrics, and implementation details.

\subsubsection{Datasets}

\begin{table*}[!t]
  \captionsetup{skip=5pt}
\caption{Dataset statistics in the experiments}
\footnotesize
\centering
\begin{tabularx}{\textwidth}{|l|X|X|X|X|X|X|X|X|X|X|}
\hline
\textbf{Scenarios} & \multicolumn{3}{c|}{\textbf{Exercise-aspect (Humanities)}} & \multicolumn{3}{c|}{\textbf{Exercise-aspect (Sciences)}} & \multicolumn{4}{c|}{\textbf{Student-aspect (Mathematics)}} \\
\hline
\textbf{Domains} & \multicolumn{1}{c|}{\textbf{Chinese}} & \multicolumn{1}{c|}{\textbf{History}} & \multicolumn{1}{c|}{\textbf{Geography}} & \multicolumn{1}{c|}{\textbf{Mathematics}} & \multicolumn{1}{c|}{\textbf{Physics}} & \multicolumn{1}{c|}{\textbf{Biology}} & \multicolumn{1}{c|}{\textbf{A-bin}} & \multicolumn{1}{c|}{\textbf{B-bin}} & \multicolumn{1}{c|}{\textbf{C-bin}} & \multicolumn{1}{c|}{\textbf{D-bin}} \\
\hline
Student Number & 4,021 & 4,021 & 4,021 & 4,021 & 4,021 & 4,021 & 1,758 & 984 & 824 & 455 \\ \hline
Exercise Number & 92 & 164 & 117 & 137 & 115 & 120 & 137 & 137 & 137 & 137 \\ \hline
Concept Number & 14 & 12 & 24 & 31 & 34 & 16 & 31 & 31 & 31 & 31 \\ \hline
Total Interactions & 263,485 & 583,334 & 381,772 & 435,797 & 387,535 & 400,858 & 197,048 & 103,852 & 86,940 & 47,957 \\ \hline
Interactions Per Student & 66 & 145 & 95 & 108 & 96 & 100 & 112 & 106 & 106 & 105 \\ \hline
Sparsity & 0.29 & 0.12 & 0.19 & 0.21 & 0.16 & 0.17 & 0.18 & 0.23 & 0.23 & 0.23 \\ \hline
Positive\_Negative Ratio & 7.04 & 3.03 & 1.69 & 2.80 & 1.49 & 1.96 & 4.46 & 2.72 & 1.87 & 1.35\\ 
\hline
\end{tabularx}
\label{tab_slp}
\end{table*}

\begin{figure}[!t]
  \centering
  \includegraphics[scale=0.16]{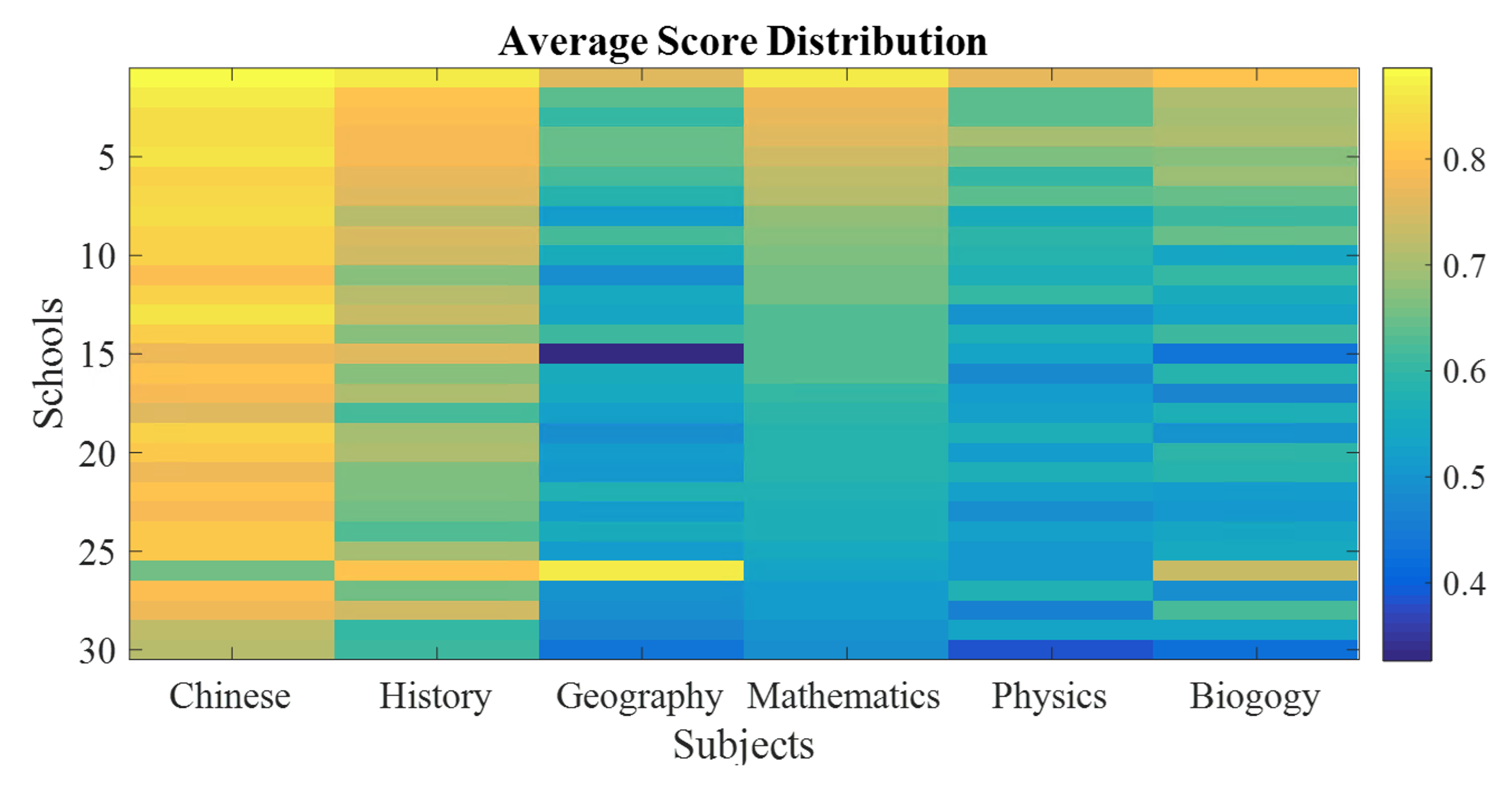}
  \caption{
  The average score distribution across different schools and subjects in the SLP dataset
  }
  \label{fig_slp}
\end{figure}

SLP \cite{lu2021slp} is a real-world educational dataset that collects students from different schools' responses to multiple subjects in K-12 education. 
The average score distribution across different schools and subjects in the SLP dataset is illustrated in Figure~\ref{fig_slp}. Research has identified variations in the interaction levels of students when answering exercises across different subjects and schools. For students within the same school, there are similar patterns in their interaction levels across different subjects. 
From the perspective of student interactions, different schools represent distinct domains in the aforementioned phenomenon. This observation precisely confirms the challenges described in Section \ref{introduction} regarding CDCD task, as similar challenges also arise from the exercise-aspect perspective.

To validate the exercise-aspect CDCD, we extracted interaction data from the SLP dataset for three humanities subjects (Chinese, History, Geography) and three science subjects (Mathematics, Physics, and Biology). For the student-aspect CDCD, we focused on interaction data from students at 30 schools in a single subject(e.g. Mathematics). We addressed the challenge of varying average cognitive levels by categorizing the schools into four bins (A, B, C, and D) based on average scores. The dataset statistics for these scenarios are presented in Table \ref{tab_slp}.

\subsubsection{Baselines}
We utilized four widely recognized CD models as the backbone diagnostic models: IRT \cite{embretson2013irt}, MIRT \cite{reckase2009mirt}, NeuralCD \cite{wang2022neuralcd}, and KSCD \cite{ma2022kscd}.
We applied our framework to these models, denoted as [Backbone]-Ours. If the prompt-to-representation mapping (Section \ref{sec_improvement}) is incorporated, the model is denoted as [Backbone]-Ours+.
The original backbone versions without cross-domain prompt transfer are used as baselines, denoted as [Backbone]-Origin. Additionally, we included state-of-the-art cross-domain cognitive diagnosis models, TechCD \cite{gao2023leveraging}, ZeroCD \cite{gao2023zero}, and CCLMF \cite{PTADisc} as baselines, referred to as [Backbone]-Tech, [Backbone]-Zero, and [Backbone]-CCLMF, respectively. For student-aspect, TechCD and ZeroCD utilize source domain representations as initial representations for the target domains.

\begin{itemize}
    \item \textbf{Origin}: The backbones without cross-domain prompt and parameter transfer.
    \item \textbf{Tech}: Apply TechCD \cite{gao2023leveraging} to the backbone. It requires the relations between knowledge concepts, constructed using a statistical method proposed in RCD \cite{gao2021rcd}.
    \item \textbf{Zero}: Apply ZeroCD \cite{gao2023zero} to the backbone.
    \item \textbf{CCLMF}: Apply CCLMF \cite{PTADisc} to the backbone.
    \item \textbf{Ours}: Apply the proposed~\shortname~to the backbone.
    \item \textbf{Ours}+: Using the prompt-to-representation mapping strategy on the basis of Ours.
\end{itemize}

\subsubsection{Metrics}
\label{sec_metrics}
Cognitive diagnosis is to assess students’ proficiency. However, since proficiency is an unobservable variable, researchers typically predict students’ future responses (correct or incorrect) and evaluate the model’s performance using the accuracy of these predictions. 
Therefore, we use classification evaluation metrics, namely AUC, ACC, RMSE, and F1.

\subsubsection{Implementation Details}
\label{exp_detail}
For both exercise- and student-aspect CDCD, we adopted a similar approach to obtain the data. In the exercise-aspect CDCD, we conducted separate diagnoses for the humanities and sciences, using any two subjects within each category as source domains and the remaining subject as the target domain. For the student-aspect CDCD, students from any three bins are treated as source domains, with the remaining bin as the target domain. In both cases, 20\% of the interaction records in the target domain are randomly selected for fine-tuning, while the remaining records are used for testing.
We determined the appropriate prompt dimensions for the backbone. Specifically, we set the dimensions for IRT and MIRT to 5 and 10, respectively, while NeuralCD and KSCD were set to 20.

\subsection{Overall Comparison (RQ1)}
\label{sec_comparison}
We compare the overall performance of the models under both the student- and exercise-aspect CDCD.

\begin{table*}[!t]
  \captionsetup{skip=5pt}
\caption{Comparison results in exercise-aspect CDCD scenarios}
\footnotesize
\centering
\begin{tabularx}{\textwidth}{|l|XXXX|XXXX|XXXX|}
\hline
\textbf{Target} & \multicolumn{4}{c|}{\textbf{Biology}} & \multicolumn{4}{c|}{\textbf{Mathematics}} & \multicolumn{4}{c|}{\textbf{Physics}} \\
\hline
\textbf{Metrics} & AUC & ACC & RMSE & F1 & AUC & ACC & RMSE & F1 & AUC & ACC & RMSE & F1 \\
\hline
IRT-Origin & 0.667 & 0.652 & 0.466 & 0.738 & 0.699 & 0.725 & 0.434 & 0.814 & 0.761 & 0.706 & 0.442 & 0.757 \\
IRT-Tech & 0.779 & 0.729 & 0.421 & 0.800 & 0.861 & 0.819 & 0.357 & 0.885 & 0.845 & 0.767 & 0.397 & 0.808 \\
IRT-Zero & 0.721 & 0.696 & 0.460 & 0.809 & 0.809 & 0.788 & 0.414 & 0.871 & 0.805 & 0.736 & 0.421 & 0.799 \\
IRT-CCLMF  & 0.775 & 0.723 & 0.425 & 0.808 & 0.870 & 0.817 & 0.353 & 0.884 & 0.850 & 0.761 & 0.403 & 0.813 \\
IRT-Ours & 0.798 & 0.742 & \textbf{0.411} & 0.815 & 0.874 & 0.826 & 0.347 & 0.886 & 0.858 & 0.776 & 0.387 & 0.818 \\
IRT-Ours+ & \textbf{0.799} & \textbf{0.743} & \textbf{0.411} & \textbf{0.816} & \textbf{0.880} & \textbf{0.832} & \textbf{0.342} & \textbf{0.890} & \textbf{0.864} & \textbf{0.781} & \textbf{0.384} & \textbf{0.823} \\ \hline
MIRT-Origin & 0.672 & 0.669 & 0.459 & 0.763 & 0.712 & 0.746 & 0.421 & 0.835 & 0.771 & 0.717 & 0.440 & 0.774 \\
MIRT-Tech & 0.779 & 0.727 & 0.422 & 0.797 & 0.864 & 0.817 & 0.355 & 0.878 & 0.847 & 0.766 & 0.395 & 0.807 \\
MIRT-Zero & 0.723 & 0.706 & 0.439 & 0.799 & 0.786 & 0.786 & 0.395 & 0.867 & 0.802 & 0.734 & 0.420 & 0.781 \\
MIRT-CCLMF & 0.771 & 0.726 & 0.426 & 0.805 & 0.867 & 0.810 & 0.368 & 0.879 & 0.843 & 0.768 & 0.398 & 0.813 \\
MIRT-Ours & 0.793 & 0.738 & 0.415 & 0.816 & 0.881 & 0.834 & 0.347 & \textbf{0.893} & 0.855 & 0.775 & 0.398 & \textbf{0.822} \\
MIRT-Ours+ & \textbf{0.801} & \textbf{0.743} & \textbf{0.411} & \textbf{0.809} & \textbf{0.886} & \textbf{0.838} & \textbf{0.343} & \textbf{0.893} & \textbf{0.865} & \textbf{0.785} & \textbf{0.389} & 0.821 \\ \hline
NCDM-Origin & 0.706 & 0.655 & 0.456 & 0.724 & 0.755 & 0.775 & 0.403 & 0.856 & 0.790 & 0.725 & 0.432 & 0.771 \\
NCDM-Tech & 0.780 & 0.727 & 0.421 & 0.795 & 0.865 & 0.809 & 0.360 & 0.874 & 0.797 & 0.732 & 0.423 & 0.784 \\
NCDM-Zero & 0.734 & 0.697 & 0.437 & 0.792 & 0.824 & 0.788 & 0.373 & 0.871 & 0.791 & 0.714 & 0.438 & 0.746 \\
NCDM-CCLMF & 0.765 & 0.731 & 0.424 & 0.811 & 0.844 & 0.808 & 0.363 & 0.875 & 0.839 & 0.769 & 0.403 & 0.809 \\
NCDM-Ours & 0.785 & \textbf{0.735} & \textbf{0.417} & \textbf{0.815} & 0.852 & 0.813 & 0.359 & \textbf{0.878} & 0.848 & 0.764 & 0.397 & 0.796 \\ 
NCDM-Ours+ & \textbf{0.788} & 0.731 & 0.418 & 0.812 & \textbf{0.872} & \textbf{0.816} & \textbf{0.357} & 0.874 & \textbf{0.861} & \textbf{0.782} & \textbf{0.386} & \textbf{0.820} \\ \hline
KSCD-Origin & 0.710 & 0.691 & 0.445 & 0.779 & 0.761 & 0.774 & 0.401 & 0.854 & 0.797 & 0.729 & 0.426 & 0.772 \\
KSCD-Tech & 0.778 & 0.729 & 0.422 & 0.799 & 0.859 & 0.818 & 0.356 & 0.882 & 0.842 & 0.765 & 0.398 & 0.807 \\
KSCD-Zero & 0.728 & 0.703 & 0.431 & 0.792 & 0.801 & 0.793 & 0.382 & 0.872 & 0.798 & 0.732 & 0.428 & 0.788 \\
KSCD-CCLMF & 0.782 & 0.732 & 0.420 & 0.799 & 0.861 & 0.815 & 0.357 & 0.879 & 0.843 & 0.765 & 0.397 & 0.813 \\
KSCD-Ours & 0.795 & \textbf{0.741} & \textbf{0.413} & \textbf{0.818} & 0.869 & 0.826 & \textbf{0.349} & \textbf{0.888} & \textbf{0.855} & \textbf{0.777} & \textbf{0.389} & \textbf{0.817} \\
KSCD-Ours+ & \textbf{0.796} & 0.739 & 0.414 & 0.812 & \textbf{0.870} & \textbf{0.828} & 0.350 & 0.886 & \textbf{0.855} & 0.776 & \textbf{0.389} & 0.816 \\ \hline
\hline
\textbf{Target} & \multicolumn{4}{c|}{\textbf{Chinese}} & \multicolumn{4}{c|}{\textbf{History}} & \multicolumn{4}{c|}{\textbf{Geography}} \\
\hline
\textbf{Metrics} & AUC & ACC & RMSE & F1 & AUC & ACC & RMSE & F1 & AUC & ACC & RMSE & F1 \\
\hline
IRT-Origin & 0.736 & 0.872 & 0.319 & 0.931 & 0.707 & 0.752 & 0.415 & 0.843 & 0.687 & 0.659 & 0.465 & 0.731 \\
IRT-Tech & 0.830 & 0.862 & 0.319 & 0.922 & 0.799 & 0.735 & 0.417 & 0.810 & 0.754 & 0.707 & 0.438 & 0.782 \\
IRT-Zero & 0.821 & 0.878 & 0.306 & 0.934 & 0.790 & 0.774 & 0.391 & 0.859 & 0.760 & 0.705 & 0.437 & 0.780 \\
IRT-CCLMF & 0.855 & 0.872 & 0.296 & 0.934 & 0.803 & 0.778 & 0.381 & 0.865 & 0.776 & 0.711 & 0.433 & 0.786 \\
IRT-Ours & 0.864 & \textbf{0.886} & 0.290 & 0.937 & 0.811 & 0.791 & 0.377 & 0.870 & 0.783 & 0.722 & 0.425 & 0.790 \\
IRT-Ours+ & \textbf{0.865} & \textbf{0.886} & \textbf{0.289} & \textbf{0.938} & \textbf{0.814} & \textbf{0.791} & \textbf{0.376} & \textbf{0.871} & \textbf{0.787} & \textbf{0.724} & \textbf{0.424} & \textbf{0.793} \\ \hline
MIRT-Origin & 0.744 & 0.873 & 0.319 & 0.932 & 0.719 & 0.756 & 0.410 & 0.846 & 0.689 & 0.671 & 0.461 & 0.763 \\
MIRT-Tech & 0.811 & 0.742 & 0.411 & 0.835 & 0.795 & 0.752 & 0.408 & 0.829 & 0.761 & 0.709 & 0.435 & 0.781 \\
MIRT-Zero & 0.822 & 0.879 & 0.302 & 0.935 & 0.782 & 0.776 & 0.392 & 0.856 & 0.724 & 0.662 & 0.454 & 0.724 \\
MIRT-CCLMF & 0.845 & 0.867 & 0.296 & 0.930 & 0.804 & 0.774 & 0.385 & 0.861 & 0.758 & 0.704 & 0.439 & 0.776 \\
MIRT-Ours & 0.863 & 0.885 & 0.289 & 0.937 & 0.818 & 0.791 & 0.376 & \textbf{0.871} & 0.778 & 0.714 & 0.431 & \textbf{0.793} \\
MIRT-Ours+ & \textbf{0.866} & \textbf{0.888} & \textbf{0.288} & \textbf{0.937} & \textbf{0.822} & \textbf{0.794} & \textbf{0.374} & 0.870 & \textbf{0.792} & \textbf{0.728} & \textbf{0.422} & 0.787 \\ \hline
NCDM-Origin & 0.782 & 0.851 & 0.323 & 0.916 & 0.742 & 0.740 & 0.412 & 0.826 & 0.717 & 0.679 & 0.453 & 0.752 \\
NCDM-Tech & 0.809 & 0.875 & 0.308 & 0.933 & 0.740 & 0.769 & 0.405 & 0.866 & 0.721 & 0.678 & 0.451 & 0.749 \\
NCDM-Zero & 0.805 & 0.876 & 0.306 & 0.932 & 0.742 & 0.761 & 0.407 & 0.863 & 0.728 & 0.687 & 0.445 & 0.755 \\
NCDM-CCLMF & 0.838 & 0.865 & 0.305 & 0.930 & 0.789 & 0.772 & 0.397 & 0.850 & 0.766 & 0.711 & 0.438 & 0.775 \\
NCDM-Ours & 0.843 & \textbf{0.881} & 0.297 & \textbf{0.934} & 0.793 & 0.778 & 0.389 & 0.856 & 0.774 & 0.717 & 0.430 & 0.780 \\
NCDM-Ours+ & \textbf{0.852} & 0.878 & \textbf{0.296} & 0.931 & \textbf{0.807} & \textbf{0.788} & \textbf{0.379} & \textbf{0.868} & \textbf{0.786} & \textbf{0.720} & \textbf{0.426} & \textbf{0.783} \\ \hline
KSCD-Origin & 0.787 & 0.869 & 0.318 & 0.929 & 0.744 & 0.771 & 0.402 & 0.857 & 0.722 & 0.688 & 0.448 & 0.771 \\
KSCD-Tech & 0.829 & 0.797 & 0.371 & 0.875 & 0.798 & 0.749 & 0.410 & 0.825 & 0.756 & 0.706 & 0.438 & 0.772 \\
KSCD-Zero & 0.825 & 0.851 & 0.328 & 0.914 & 0.789 & 0.769 & 0.406 & 0.853 & 0.749 & 0.702 & 0.443 & 0.778 \\
KSCD-CCLMF & 0.842 & 0.875 & 0.305 & 0.925 & 0.793 & 0.776 & 0.388 & 0.856 & 0.772 & 0.711 & 0.432 & 0.787 \\
KSCD-Ours & \textbf{0.854} & 0.883 & \textbf{0.292} & \textbf{0.936} & \textbf{0.807} & \textbf{0.789} & \textbf{0.380} & 0.868 & 0.785 & \textbf{0.723} & \textbf{0.425} & 0.794 \\
KSCD-Ours+ & \textbf{0.854} & \textbf{0.884} & \textbf{0.292} & \textbf{0.936} & 0.804 & \textbf{0.789} & \textbf{0.380} & \textbf{0.869} & \textbf{0.787} & 0.719 & 0.426 & \textbf{0.797} \\ \hline
\end{tabularx}
\label{tab_performance1}
\end{table*}

\textbf{Exercise-Aspect CDCD.} Table \ref{tab_performance1} presents the performance under the exercise-aspect CDCD. The cross-subject scenario divides six subjects into two categories: humanities and sciences, with three subjects in each category. Two subjects are treated as source domains, and one as the target domain. Across all target domains, the proposed ~\shortname~ consistently outperforms state-of-the-art baselines. Specifically, when comparing the baselines and our framework across the IRT, MIRT, NeuralCD, and KSCD backbones, our method demonstrates significant improvement in the AUC metric, with an average increase of up to nearly 20\% compared to the Origin version, especially in the Biology ($0.667 \rightarrow \boldsymbol{0.798}$ in IRT) and Chinese subjects ($0.736 \rightarrow \boldsymbol{0.864}$ in IRT). Additionally, as a unified framework for cross-domain scenarios, our approach demonstrates notable improvements over TechCD \cite{gao2023leveraging}, ZeroCD \cite{gao2023zero}, and CCLMF \cite{PTADisc}. Similarly, the RMSE metric shows a significant reduction in all scenarios. The Ours+ version, which incorporates an additional adaptation strategy, shows consistent improvements over the Ours version in most test scenarios, further validating the robustness of our approach.


\textbf{Student-Aspect CDCD.} Table \ref{tab_performance2} showcases the performance under the student-aspect CDCD. In this setup, schools are categorized into four bins (A, B, C, D) based on their average scores, with three bins designated as source domains and the remaining as the target domain. Specifically, in the A-bin target domain, the NeuralCD-Ours shows 27.9\% improvement ($0.687 \rightarrow \boldsymbol{0.879}$) in the AUC metric relative to the origin, while KSCD-Ours achieves 14.5\% increase ($0.764 \rightarrow \boldsymbol{0.875}$). Through the aforementioned experiments, we observed similar patterns across other metrics when applying PromptCD in various scenarios.

Comparing the two tables, we see that our proposed framework shows a significant improvement over TechCD and ZeroCD in Table \ref{tab_performance2} than in Table \ref{tab_performance1}, primarily because the baseline algorithms are not specifically designed to address student-aspect CDCD.

\begin{table*}[!t]
  \captionsetup{skip=5pt}
\caption{Comparison results in student-aspect CDCD scenarios}
\footnotesize
\centering
\begin{tabularx}{\textwidth}{|l|XXXX|XXXX|XXXX|XXXX|}
\hline
\textbf{Target} & \multicolumn{4}{c|}{\textbf{A-bin}} & \multicolumn{4}{c|}{\textbf{B-bin}} & \multicolumn{4}{c|}{\textbf{C-bin}} & \multicolumn{4}{c|}{\textbf{D-bin}} \\
\hline
\textbf{Metrics} & AUC & ACC & RMSE & F1 & AUC & ACC & RMSE & F1 & AUC & ACC & RMSE & F1 & AUC & ACC & RMSE & F1\\
\hline
IRT-Origin & 0.670 & 0.800 & 0.393 & 0.884 & 0.548 & 0.728 & 0.451 & 0.841 & 0.678 & 0.679 & 0.477 & 0.777 & 0.519 & 0.537 & 0.512 & 0.630 \\
IRT-Tech & 0.821 & 0.847 & 0.336 & 0.912 & 0.837 & 0.811 & 0.365 & 0.877 & 0.769 & 0.741 & 0.425 & 0.830 & 0.809 & 0.733 & 0.419 & 0.767 \\
IRT-Zero & 0.855 & 0.854 & 0.324 & 0.917 & 0.854 & 0.823 & 0.360 & 0.884 & 0.847 & 0.784 & 0.390 & 0.835 & 0.831 & 0.753 & 0.406 & 0.792 \\
IRT-CCLMF  & 0.851 & 0.854 & 0.321 & 0.920 & 0.853 & 0.822 & 0.362 & 0.883 & 0.855 & 0.790 & 0.384 & 0.847 & 0.839 & 0.757 & 0.403 & 0.799 \\
IRT-Ours & 0.871 & 0.867 & 0.316 & 0.922 & 0.870 & 0.826 & 0.351 & 0.884 & 0.877 & 0.806 & 0.368 & \textbf{0.858} & 0.857 & \textbf{0.766} & 0.397 & \textbf{0.814} \\
IRT-Ours+ & \textbf{0.881} & \textbf{0.872} & \textbf{0.308} & \textbf{0.925} & \textbf{0.881} & \textbf{0.834} & \textbf{0.344} & \textbf{0.89} & \textbf{0.881} & \textbf{0.811} & \textbf{0.363} & \textbf{0.858} & \textbf{0.858} & 0.764 & \textbf{0.395} & 0.813 \\ \hline
MIRT-Origin & 0.718 & 0.820 & 0.372 & 0.896 & 0.742 & 0.762 & 0.413 & 0.848 & 0.715 & 0.706 & 0.461 & 0.798 & 0.729 & 0.682 & 0.470 & 0.749 \\
MIRT-Tech & 0.820 & 0.845 & 0.339 & 0.912 & 0.840 & 0.811 & 0.361 & 0.878 & 0.805 & 0.760 & 0.403 & 0.831 & 0.817 & 0.739 & 0.414 & 0.780 \\
MIRT-Zero & 0.841 & 0.841 & 0.347 & 0.902 & 0.849 & 0.814 & 0.366 & 0.880 & 0.845 & 0.799 & 0.367 & 0.854 & 0.804 & 0.735 & 0.433 & 0.754 \\
MIRT-CCLMF & 0.834 & 0.849 & 0.331 & 0.911 & 0.842 & 0.814 & 0.368 & 0.881 & 0.842 & 0.770 & 0.398 & 0.841 & 0.814 & 0.742 & 0.428 & 0.762 \\
MIRT-Ours & 0.861 & 0.859 & 0.324 & 0.917 & 0.863 & 0.816 & 0.365 & 0.883 & 0.862 & 0.778 & 0.395 & 0.844 & 0.844 & 0.766 & 0.409 & 0.806 \\
MIRT-Ours+ & \textbf{0.886} & \textbf{0.872} & \textbf{0.311} & \textbf{0.923} & \textbf{0.886} & \textbf{0.836} & \textbf{0.347} & \textbf{0.891} & \textbf{0.881} & \textbf{0.807} & \textbf{0.375} & \textbf{0.858} & \textbf{0.859} & \textbf{0.778} & \textbf{0.398} & \textbf{0.813} \\ \hline
NCDM-Origin & 0.687 & 0.809 & 0.387 & 0.894 & 0.693 & 0.743 & 0.465 & 0.847 & 0.709 & 0.643 & 0.485 & 0.782 & 0.533 & 0.574 & 0.653 & 0.729 \\
NCDM-Tech & 0.818 & 0.845 & 0.340 & 0.910 & 0.838 & 0.805 & 0.364 & 0.873 & 0.816 & 0.766 & 0.401 & 0.835 & 0.796 & 0.697 & 0.429 & 0.714 \\
NCDM-Zero & 0.761 & 0.800 & 0.376 & 0.877 & 0.798 & 0.712 & 0.430 & 0.783 & 0.754 & 0.724 & 0.430 & 0.804 & 0.797 & 0.706 & 0.455 & 0.718 \\
NCDM-CCLMF & 0.844 & 0.851 & 0.332 & 0.911 & 0.846 & 0.813 & 0.368 & 0.871 & 0.838 & 0.775 & 0.396 & 0.820 & 0.813 & 0.734 & 0.414 & 0.757 \\
NCDM-Ours & \textbf{0.879} & \textbf{0.870} & \textbf{0.308} & \textbf{0.924} & 0.865 & 0.820 & 0.357 & 0.877 & 0.856 & 0.780 & 0.390 & 0.822 & 0.837 & 0.751 & \textbf{0.408} & 0.765 \\
NCDM-Ours+ & 0.878 & 0.865 & 0.319 & 0.920 & \textbf{0.878} & \textbf{0.833} & \textbf{0.352} & \textbf{0.891} & \textbf{0.864} & \textbf{0.796} & \textbf{0.380} & \textbf{0.841} & \textbf{0.840} & \textbf{0.752} & 0.409 & \textbf{0.785} \\ \hline
KSCD-Origin & 0.764 & 0.831 & 0.368 & 0.901 & 0.756 & 0.775 & 0.414 & 0.860 & 0.766 & 0.726 & 0.448 & 0.807 & 0.769 & 0.706 & 0.449 & 0.737 \\
KSCD-Tech & 0.809 & 0.848 & 0.338 & 0.913 & 0.839 & 0.809 & 0.366 & 0.884 & 0.780 & 0.756 & 0.412 & 0.830 & 0.794 & 0.725 & 0.425 & 0.772 \\
KSCD-Zero & 0.778 & 0.806 & 0.346 & 0.877 & 0.808 & 0.798 & 0.390 & 0.879 & 0.785 & 0.763 & 0.408 & 0.839 & 0.797 & 0.737 & 0.423 & 0.782 \\
KSCD-CCLMF & 0.854 & 0.855 & 0.327 & 0.915 & 0.853 & 0.822 & 0.358 & 0.886 & 0.853 & 0.794 & 0.385 & 0.846 & 0.818 & 0.759 & 0.418 & 0.786 \\
KSCD-Ours & 0.875 & 0.865 & 0.312 & 0.921 & 0.876 & 0.834 & 0.346 & \textbf{0.891} & \textbf{0.871} & \textbf{0.806} & \textbf{0.371} & \textbf{0.854} & 0.844 & 0.768 & \textbf{0.403} & \textbf{0.805} \\
KSCD-Ours+ & \textbf{0.880} & \textbf{0.868} & \textbf{0.310} & \textbf{0.923} & \textbf{0.878} & \textbf{0.835} & \textbf{0.345} & \textbf{0.891} & 0.870 & 0.804 & \textbf{0.371} & 0.851 & \textbf{0.846} & \textbf{0.771} & 0.404 & 0.795 \\ \hline

\end{tabularx}
\label{tab_performance2}
\end{table*}

\textbf{Significance Analysis of Model Performance.}
We conducted Nemenyi tests on various baseline models used in different scenarios to report statistical significance for metrics such as AUC, ACC, RMSE and F1.

\begin{figure}[!t]
  \centering
  \includegraphics[width=1\columnwidth]{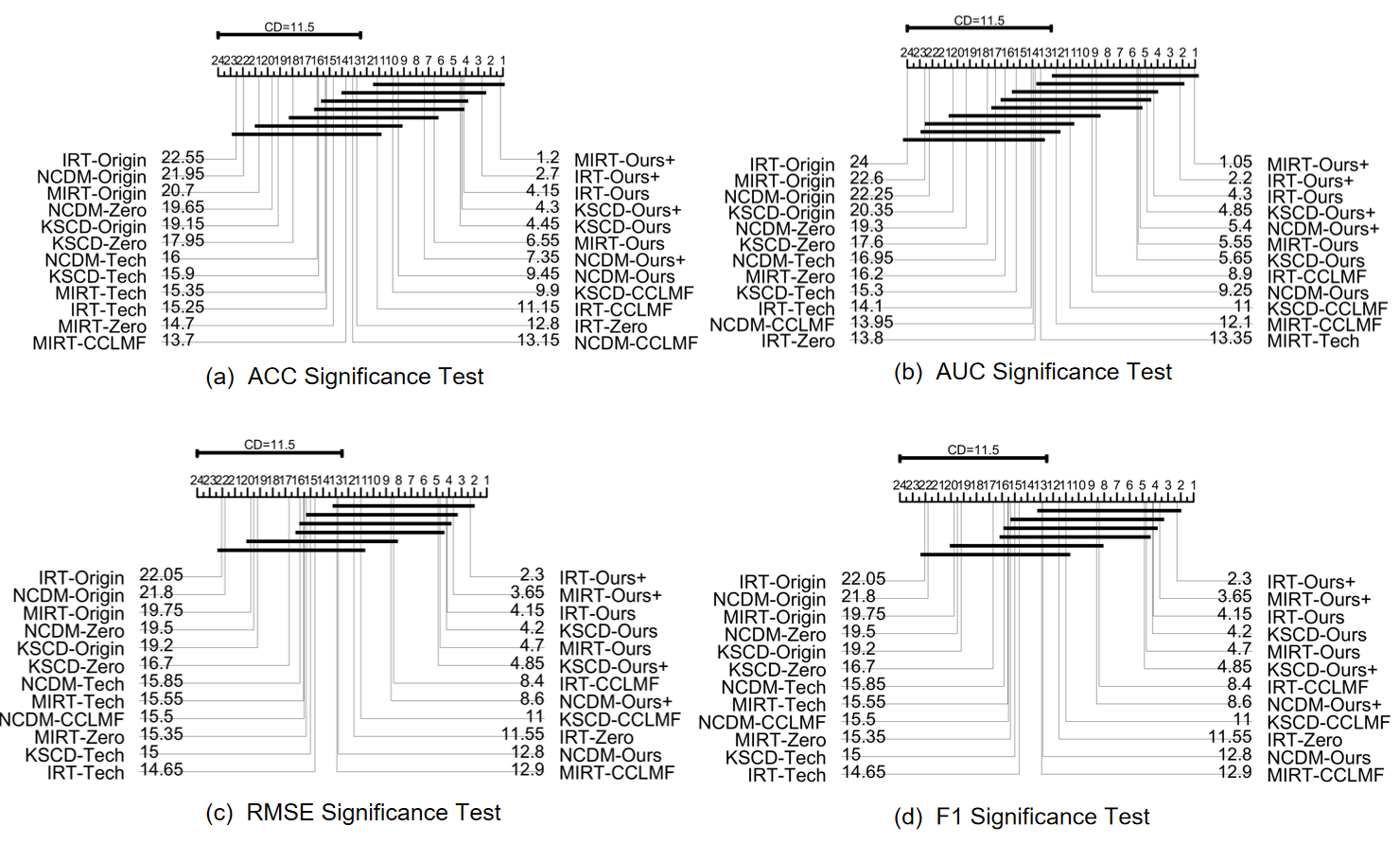}
  \caption{
Visualization of Significance Test Results for Evaluation Metrics
  }
  \label{fig_significanse}
\end{figure}

As shown in Figure \ref{fig_significanse}, the PromptCD model (especially the "Ours+" version) significantly outperforms other baseline models across multiple domains. This further demonstrates the effectiveness and robustness of cross-domain prompt transfer methods in cognitive diagnosis tasks.

\subsection{Detailed Analysis (RQ2)}
To address the role of~\shortname~in key aspects, we analyze the effects of different fine-tuning ratios, the choice of source domains, and prompt dimensions in horizontal concatenation.

\begin{figure*}[!t]
  \centering
  \includegraphics[scale=0.3]{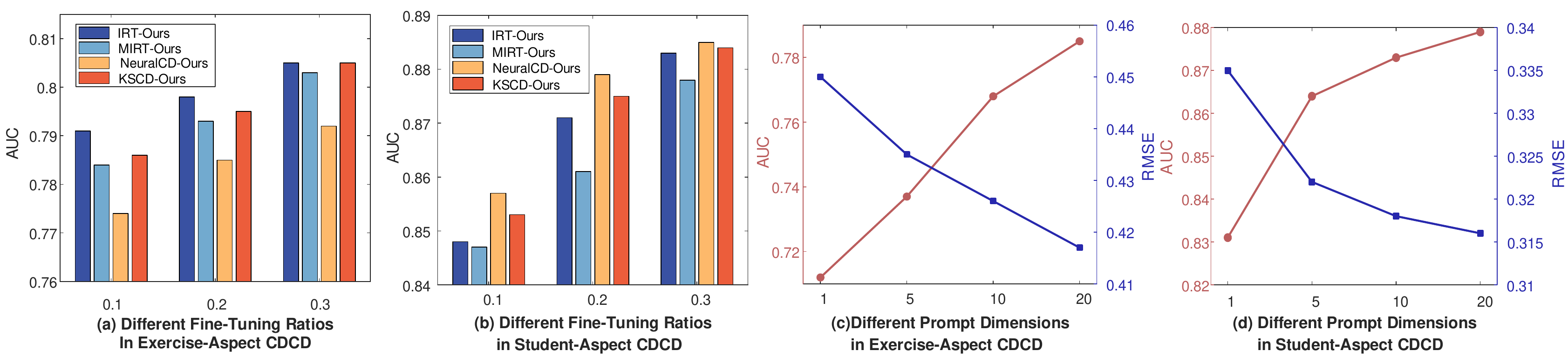}
  \caption{
Performance comparisons with (a-b) different tuning ratios and (c-d) different prompt dimensions
  }
  \label{fig_para}
\end{figure*}

\textbf{Different Fine-tuning Ratios.}
We examine the performance of the~\shortname~across varying proportions of fine-tuning data, addressing the challenge of data sparsity in cross-domain knowledge adaptation. Specifically, we evaluate the model's performance with fine-tuning data proportions of 0.1, 0.2, and 0.3 on both the exercise-aspect CDCD for the biology target domain and the student-aspect for the A-bin target domain, shown in Figure. \ref{fig_para} (a) and (b).
In most cases, the performance improves as the proportion increases, even at a small proportion of 0.1, the models can achieve satisfactory performance in both student-aspect and exercise-aspect scenarios by using ~\shortname~. 

\begin{table*}[!t]
\caption{Performance comparisons with different source domains in exercise and student-aspect scenarios}
\footnotesize
\centering
\begin{tabularx}{\textwidth}{|l|l|XXXX|l|XXXX|}
\hline
\multirow{2}{*}{\textbf{Backbone}} & \multirow{2}{*}{\textbf{Source Domain}} & \multicolumn{4}{c|}{\textbf{Biology}} & \multirow{2}{*}{\textbf{Source Domain}} & \multicolumn{4}{c|}{\textbf{A-bin}} \\

 &  & AUC & ACC & RMSE & F1 &  & AUC & ACC & RMSE & F1 \\
 \hline
\multirow{3}{*}{IRT} & Mathematics & 0.781 & 0.729 & 0.414 & 0.814 & B & 0.860 & 0.861 & 0.321 & 0.919 \\
 & Physics & 0.782 & 0.729 & 0.413 & 0.814 & B+C & 0.870 & 0.865 & \textbf{0.314} & 0.921 \\
 & Mathematics+Physics & \textbf{0.798} & \textbf{0.742} & \textbf{0.411} & \textbf{0.815} & B+C+D & \textbf{0.871} & \textbf{0.867} & 0.316 & \textbf{0.922} \\
 \hline
\multirow{3}{*}{MIRT}  & Mathematics & 0.781 & 0.730 & 0.421 & 0.814 & B & 0.853 & 0.856 & 0.324 & 0.917 \\
 & Physics & 0.784 & 0.732 & 0.419 & 0.814 & B+C & \textbf{0.862} & \textbf{0.861} & \textbf{0.320} & \textbf{0.919} \\
 & Mathematics+Physics & \textbf{0.793} & \textbf{0.738} & \textbf{0.415} & \textbf{0.816} & B+C+D & 0.861 & 0.859 & 0.321 & \textbf{0.919} \\
 \hline
\multirow{3}{*}{NeuralCD} & Mathematics & 0.774 & 0.726 & 0.423 & 0.796 & B & 0.865& 0.862 & 0.318 & 0.920 \\
 & Physics & 0.773 & 0.726 & 0.423 & 0.800 & B+C & 0.874 & 0.868 & 0.311 & 0.922 \\
 & Mathematics+Physics & \textbf{0.785} & \textbf{0.735} & \textbf{0.417} & \textbf{0.815} & B+C+D & \textbf{0.879} & \textbf{0.870} & \textbf{0.308} & \textbf{0.924}\\
 \hline
 \multirow{3}{*}{KSCD} & Mathematics & 0.782 & 0.735 & 0.418 & 0.812 & B & 0.868 & 0.864 & 0.316 & 0.921 \\
 & Physics & 0.781 & 0.734 & 0.418 & 0.815 & B+C & 0.875 & \textbf{0.868} & \textbf{0.311} & \textbf{0.923} \\
 & Mathematics+Physics & \textbf{0.795} & \textbf{0.741} & \textbf{0.413} & \textbf{0.818} & B+C+D & \textbf{0.875} & 0.865 & 0.312 & 0.921\\
 \hline
\end{tabularx}
\label{tab_parameter1}
\end{table*}

\textbf{Different Prompt Dimensions.}
We demonstrate the influence of prompt dimensionality on ~\shortname. We provide results using the NeuralCD, which employs horizontal concatenation, as an example.
Specifically, we set the prompt dimensionality to 1, 5, 10, or 20. The comparison results are shown in Figure. \ref{fig_para} (c) and (d). Performance is lowest when the prompt dimension is 1. As the prompt dimension increases, performance improves, reflecting the enhanced information capacity of the prompts and their effectiveness in improving model performance.

\textbf{Various Source Domains.}
We evaluate the impact of various source domains ~\shortname. Specifically, for the exercise-aspect scenario with biology as the target domain, We consider three scenarios: using mathematics, physics, or both as source domains, respectively. Similarly, for the student-aspect scenario with A-bin as the target domain, we consider three types of combinations of source domains, which are illustrated in Table \ref{tab_parameter1}. The performance achieved using data from two source domains is superior to that obtained from a single source domain, which indicates that ~\shortname~ can learn common knowledge across multiple source domains.

\textbf{Various Cross-Domain Types.} Table \ref{tab_performance_added} presents the experimental results of the model in more scenarios, including from humanities to sciences and from sciences to humanities. The detailed experimental setups are as follows: \begin{itemize} \item \textbf{Sciences-Humanities}: Source: Biology, Mathematics $\rightarrow$ Target: Geography \item \textbf{Humanities-Humanities}: Source: Chinese, History $\rightarrow$ Target: Geography \item \textbf{Humanities-Sciences}: Source: Chinese, History $\rightarrow$ Target: Physics \item \textbf{Sciences-Sciences}: Source: Biology, Mathematics $\rightarrow$ Target: Physics \end{itemize}
Our results indicate that in both cases, PromptCD outperforms the comparison algorithms. Interestingly, its performance shows a slight decline compared to the ``humanities to sciences'' and ``sciences to sciences'' scenarios, which aligns with empirical expectations. Specifically, the transfer of student states is more effective between disciplines with similar characteristics, whereas greater differences between disciplines result in more significant deviations in the transferred student states.

\begin{table*}[!t]
  \captionsetup{skip=5pt}
\caption{Comparison results in added CDCD scenarios}
\footnotesize
\centering
\begin{tabularx}{\textwidth}{|l|XXXX|XXXX|XXXX|XXXX|}
\hline
\textbf{Scenarios} & \multicolumn{4}{c|}{\textbf{Sciences-Humanities}} & \multicolumn{4}{c|}{\textbf{Humanities-Humanities}} & \multicolumn{4}{c|}{\textbf{Humanities-Sciences}} & \multicolumn{4}{c|}{\textbf{Sciences-Sciences}} \\
\hline
\textbf{Metrics} & AUC & ACC & RMSE & F1 & AUC & ACC & RMSE & F1& AUC & ACC & RMSE & F1 & AUC & ACC & RMSE & F1\\ \hline 
IRT-Origin & 0.677 & 0.650 & 0.472 & 0.721 & 0.687 & 0.659 & 0.465 & 0.731 & 0.747 & 0.686 & 0.460 & 0.731 & 0.761 & 0.706 & 0.442 & 0.757\\ 
IRT-Tech & 0.757 & 0.705 & 0.438 & 0.784 & 0.754 & 0.707 & 0.438 & 0.782 & 0.829 & 0.753 & 0.407 & 0.791 & 0.845 & 0.767 & 0.397 & 0.808\\ 
IRT-Zero & 0.734 & 0.693 & 0.446 & 0.776 & 0.760 & 0.705 & 0.437 & 0.780 & 0.804 & 0.730 & 0.424 & 0.766 & 0.805 & 0.736 & 0.421 & 0.799\\ 
IRT-CCLMF & 0.775 & 0.715 & 0.427 & 0.791 & 0.776 & 0.711 & 0.433 & 0.786 & 0.804 & 0.730 & 0.424 & 0.766 & 0.805 & 0.736 & 0.421 & 0.799\\  
IRT-Ours & 0.791 & 0.726 & 0.422 & \textbf{0.795} & 0.783 & 0.722 & 0.425 & 0.790 & \textbf{0.854} & 0.773 & \textbf{0.390} & \textbf{0.815} & 0.858 & 0.776 & 0.387 & 0.818\\  
IRT-Ours+ & \textbf{0.792} & \textbf{0.727} & \textbf{0.421} & 0.792 & \textbf{0.787} & \textbf{0.724} & \textbf{0.424} & \textbf{0.793} & \textbf{0.854} & \textbf{0.774} & \textbf{0.390} & 0.814 & \textbf{0.864} & \textbf{0.781} & \textbf{0.384} & \textbf{0.823}\\ \hline 
MIRT-Origin & 0.695 & 0.669 & 0.459 & 0.773 & 0.689 & 0.671 & 0.461 & 0.763 & 0.775 & 0.718 & 0.439 & 0.779 & 0.771 & 0.717 & 0.440 & 0.774\\ 
MIRT-Tech & 0.765 & 0.709 & 0.435 & 0.787 & 0.761 & 0.709 & 0.435 & 0.781 & 0.828 & 0.744 & 0.416 & 0.807 & 0.847 & 0.766 & 0.395 & 0.807\\ 
MIRT-Zero & 0.722 & 0.686 & 0.450 & 0.763 & 0.724 & 0.662 & 0.454 & 0.724 & 0.794 & 0.721 & 0.428 & 0.757 & 0.802 & 0.734 & 0.420 & 0.781\\ 
MIRT-CCLMF & 0.767 & 0.710 & 0.434 & 0.775 & 0.758 & 0.704 & 0.439 & 0.776 & 0.838 & 0.756 & 0.399 & 0.810 & 0.843 & 0.768 & 0.398 & 0.813\\ 
MIRT-Ours & 0.786 & 0.720 & 0.428 & \textbf{0.794} & 0.778 & 0.714 & 0.431 & \textbf{0.793} & 0.843 & 0.764 & 0.405 & \textbf{0.813} & 0.855 & 0.775 & 0.398 & \textbf{0.822}\\ 
MIRT-Ours+ & \textbf{0.791} & \textbf{0.728} & \textbf{0.423} & 0.789 & \textbf{0.792} & \textbf{0.728} & \textbf{0.422} & 0.787 & \textbf{0.852} & \textbf{0.771} & \textbf{0.397} & 0.806 & \textbf{0.865} & \textbf{0.785} & \textbf{0.389} & 0.821\\ \hline
NCDM-Origin & 0.714 & 0.683 & 0.460 & 0.765 & 0.717 & 0.679 & 0.453 & 0.752 & 0.782 & 0.721 & 0.435 & 0.764 & 0.790 & 0.725 & 0.432 & 0.771\\ 
NCDM-Tech & 0.771 & 0.715 & 0.431 & 0.788 & 0.721 & 0.678 & 0.451 & 0.749 & 0.821 & 0.743 & 0.414 & 0.799 & 0.797 & 0.732 & 0.423 & 0.784\\ 
NCDM-Zero & 0.718 & 0.688 & 0.451 & 0.777 & 0.728 & 0.687 & 0.445 & 0.755 & 0.798 & 0.729 & 0.426 & 0.787 & 0.791 & 0.714 & 0.438 & 0.746\\ 
NCDM-CCLMF & 0.769 & 0.715 & 0.435 & 0.781 & 0.766 & 0.711 & 0.438 & 0.775 & 0.831 & 0.754 & 0.408 & 0.791 & 0.839 & 0.769 & 0.403 & 0.809\\ 
NCDM-Ours & 0.779 & 0.720 & 0.427 & \textbf{0.784} & 0.774 & 0.717 & 0.430 & 0.780 & 0.832 & 0.757 & 0.407 & 0.806 & 0.848 & 0.764 & 0.397 & 0.796 \\ 
NCDM-Ours+ & \textbf{0.786} & \textbf{0.722} & \textbf{0.424} & 0.783 & \textbf{0.786} & \textbf{0.720} & \textbf{0.426} & \textbf{0.783} & \textbf{0.848} & \textbf{0.769} & \textbf{0.397} & \textbf{0.814} & \textbf{0.861} & \textbf{0.782} & \textbf{0.386} & \textbf{0.820}\\ \hline 
KSCD-Origin & 0.722 & 0.687 & 0.448 & 0.770 & 0.722 & 0.688 & 0.448 & 0.771 & 0.798 & 0.728 & 0.426 & 0.771 & 0.797 & 0.729 & 0.426 & 0.772\\ 
KSCD-Tech & 0.761 & 0.709 & 0.435 & 0.785 & 0.756 & 0.706 & 0.438 & 0.772 & 0.826 & 0.753 & 0.409 & 0.805 & 0.842 & 0.765 & 0.398 & 0.807\\ 
KSCD-Zero & 0.734 & 0.695 & 0.442 & 0.776 & 0.749 & 0.702 & 0.443 & 0.778 & 0.809 & 0.733 & 0.420 & 0.791 & 0.798 & 0.732 & 0.428 & 0.788\\ 
KSCD-CCLMF & 0.776 & 0.719 & 0.428 & 0.796 & 0.772 & 0.711 & 0.432 & 0.787 & 0.836 & 0.761 & 0.404 & 0.806 & 0.843 & 0.765 & 0.397 & 0.813\\ 
KSCD-Ours & \textbf{0.788} & \textbf{0.726} & \textbf{0.424} & 0.791 & 0.785 & \textbf{0.723} & \textbf{0.425} & 0.794 & 0.848 & 0.769 & 0.395 & 0.804 & 0.855 & \textbf{0.777} & \textbf{0.389} & \textbf{0.817}\\ 
KSCD-Ours+ & \textbf{0.788} & 0.724 & \textbf{0.424} & \textbf{0.794} & \textbf{0.787} & 0.719 & 0.426 & \textbf{0.797} & \textbf{0.849} & \textbf{0.772} & \textbf{0.393} & \textbf{0.815} & \textbf{0.855} & 0.776 & \textbf{0.389} & \textbf{0.816}\\ \hline
\end{tabularx}
\label{tab_performance_added}
\end{table*}

\subsection{Feature Visualization (RQ3)}
In this section, we visualized the representations learned by the model in CDCD scenarios for both exercises and students to explain why the Prompts learned from the source domains are effective.

\begin{figure*}[!t]
  \centering
  \captionsetup{skip=0pt}
  \includegraphics[width=2\columnwidth]{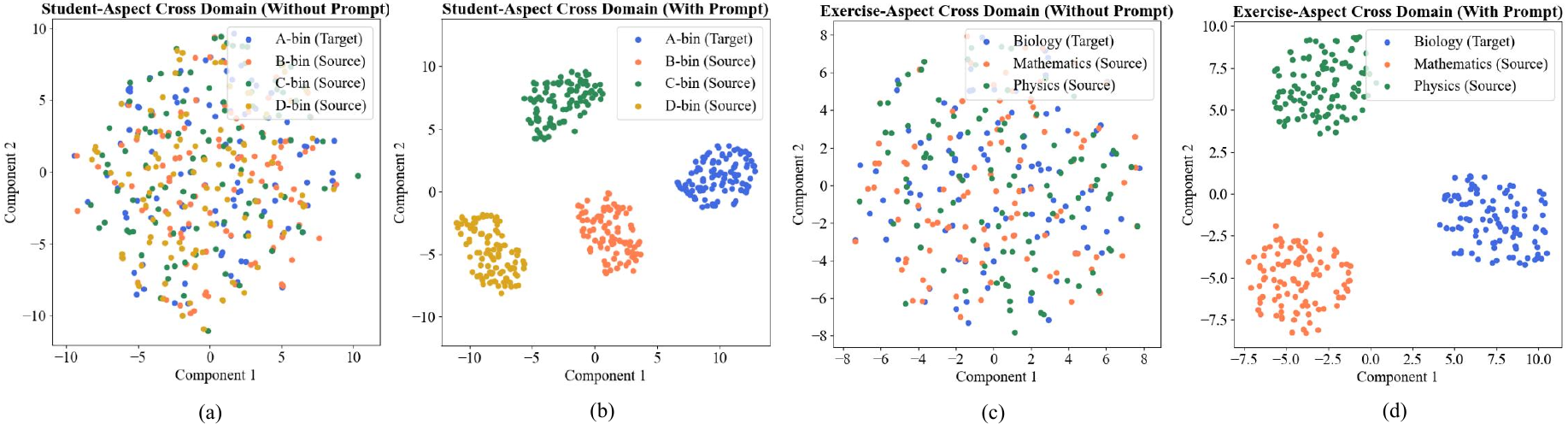}
  \caption{
Visualization of (a, c) origin representations without prompt and (b, d) our representations with prompt
  }
  \label{fig_visual}
\end{figure*}

For the student-aspect CDCD scenario, Figure. \ref{fig_visual} (a) displays the distribution of the original student representations after dimensionality reduction and reveals that the original representations of students from different bins do not display any discernible patterns. In contrast, Figure. \ref{fig_visual} (b) displays the distribution of the final representations, obtained after the operations described in Section \ref{sec:Source Prompt}. The clustering of students within the same bin suggests that our Prompts effectively capture the overall characteristics of the domain. Additionally, student representations from A-bin are more distant from those of D-bin and closer to those of B-bin, indicating that the prompts effectively distinguish between different levels of student groups. In the exercise-aspect CDCD, as shown in Figure \ref{fig_visual} (c) and (d), the transfer prompts also effectively capture both the internal characteristics of each subject and the distinctions between different subjects.

Additionally, we employed two quantitative metrics—inter-cluster distance and intra-cluster distance—to analyze the changes in student and exercise representations before and after introducing prompts. As shown in Table \ref{tab:cluster_analysis}, the inter-cluster and intra-cluster distances are significantly smaller after introducing prompts compared to before, providing a clearer demonstration of the effectiveness of our method and ensuring consistency in the presentation across both student and exercise dimensions.

\begin{table}[!t]
\captionsetup{skip=5pt}
\centering
\caption{Cluster Analysis Before and After Prompt}
\small
\begin{tabular}{|l|l|c|c|}
\hline
\textbf{Status}       & \textbf{Dimension}         & \textbf{Intra-Dist} & \textbf{Inter-Dist} \\ \hline
without Prompt         & Exercise Embedding         & 6.6352                          & 0.2911                          \\ \hline
without Prompt         & Student Embedding          & 7.8856                          & 0.6041                          \\ \hline
with Prompt          & Exercise Embedding         & 2.8690                          & 12.2945                         \\ \hline
with Prompt          & Student Embedding          & 2.9120                          & 13.0087                         \\ \hline
\end{tabular}
\label{tab:cluster_analysis}
\end{table}

\subsection{Personalized Recommendation (RQ4)}
In this section, we illustrate how the ~\shortname~ facilitates personalized learning guidance for exercise recommendations. We employ a straightforward yet effective strategy to suggest exercises related to concepts that students have not yet mastered \cite{gao2023leveraging}, ensuring they are of appropriate difficulty \cite{huang2019exploring}.
Based on a CD backbone, we first determine whether this is a student-side or exercise-side cross-domain recommendation scenario, then select the corresponding framework to be added to the CD model. 
After the model undergoes a two-stage training process, it produces a well-trained model $\mathcal{M}$. $\mathcal{M}$ can determine the student's mastery level of knowledge concepts through a diagnostic module. We select $N$ exercises associated with the knowledge concepts that the student has not yet mastered. Furthermore, we aim for the exercises to be of moderate difficulty for the student, as exercises that are too difficult or too easy may hurt their learning interest. Therefore, from this set of $K$ exercises, we ultimately choose $K$ exercises of moderate difficulty to form the recommended list for the student. 
Let's take the example of applying the proposed framework to NeuralCD \cite{wang2020ncdm} to showcase the results of the personalized learning recommendation. In the exercise-aspect CDCD, mathematics and biology are treated as the source domains, and physics is the target domain. We recommend exercises to a randomly sampled student. The recommended concept ID, exercise ID, corresponding student mastery, exercise difficulty, and true performance are detailed in Table \ref{recom}. The results demonstrate that the recommended exercises align with the requirements of practical applications, which are exercises that the student has not mastered yet and are of moderate difficulty.

\begin{table}[!t]
  \captionsetup{skip=5pt}
  \footnotesize
  \centering
  \caption{Recommending example in exercise-aspect CDCD}
  \setlength{\tabcolsep}{4pt} 
  \renewcommand{\arraystretch}{0.9} 
  \begin{tabular}{|l|l|l|l|l|l|l|l|}
    \hline
    Concept ID          & 11    & 14    & 19    & 25    & 26    & 28    & 33    \\
    \hline
    Exercise ID         & 16    & 2     & 14    & 25    & 23    & 54    & 3     \\
    \hline
    Student Mastery     & 0.491 & 0.472 & 0.484 & 0.489 & 0.490 & 0.481 & 0.485 \\
    \hline
    Exercise Difficulty & 0.550 & 0.481 & 0.520 & 0.428 & 0.530 & 0.487 & 0.469 \\
    \hline
    True Performance    & 0     & 0     & 0     & 1     & 0     & 0     & 1    \\
    \hline
  \end{tabular}
\label{recom}
\end{table}

\section{Related Work}
\subsection{Cognitive Diagnosis}
\label{sec_relatedwork_cd}
Cognitive diagnosis \cite{fcs2023new-dev,gao2021rcd,ma2022kscd} is a form of student learning modeling that plays a vital role in educational recommendation tasks \cite{liu2023meta,huang2019exploring}. Traditional cognitive diagnosis models, such as IRT \cite{embretson2013irt} and MIRT \cite{reckase2009mirt}, utilize unidimensional and multidimensional latent traits, respectively, to represent student and exercise features. The DINA model \cite{de2009dina} incorporates guessing and slipping parameters but often relies on assumptions that oversimplify student interactions. These traditional models have established a foundation for cognitive diagnosis but struggle to capture the complexities of student behavior and learning patterns. To address these limitations, various deep learning-based cognitive diagnosis models have been proposed. Wang et al. \cite{wang2022neuralcd} introduced NeuralCD, which uses neural networks to enhance both accuracy and interpretability. Many works have expanded upon NeuralCD, such as KaNCD \cite{wang2022neuralcd} and KSCD \cite{ma2022kscd}, which make full use of information from non-interactive knowledge concepts.
However, these models that assume the training and testing data are from the same distribution will experience a significant decline in performance when confronted with non-identical distributions.

\subsubsection*{Cross-Domain Cognitive Diagnosis}
The introduction of new domains in online education often leads to the unavailability of practice logs for many students, creating the CDCD issue \cite{gao2023leveraging}. Gao et al. \cite{gao2023leveraging} proposed TechCD for exercise-aspect CDCD, which uses transferable knowledge concept graphs to address the cold-start problem in new domains. This method embeds knowledge concepts and student behaviors into a graph, leveraging transferable knowledge to accurately assess cognitive abilities. ZeroCD \cite{gao2023zero} tackles the CDCD problem by utilizing early-participating student data to assess cognitive abilities with minimal data. In addition, Hu et al. \cite{PTADisc} proposed CCLMF, which leverages a meta-learner to predict network parameters and enhances model performance on target courses using knowledge from source courses. However, TechCD, ZeroCD, and CCLMF only address the exercise-aspect CDCD problem, focusing on just one aspect of the issue. In this paper, we focus on a scenario-agnostic CDCD framework, maintaining compatibility with both exercise-aspect and student-aspect scenarios.

\subsection{Prompt Learning}
Prompt learning \cite{petroni2019language} is a technique applied to pre-trained language models \cite{liu2023pre} and has demonstrated significant success in various applications, including recommendation tasks \cite{zhang2023prompt, li2023gpt4rec, hou2023large, wu2023towards}. This approach guides the model's generation process using prompts. Prompts can be either hard (discrete words) or soft (continuous learnable embeddings) \cite{li2023personalized}. Soft prompts, in particular, offer greater flexibility as they can be optimized and adjusted during training, allowing them to better adapt to specific tasks and data requirements \cite{gu2021ppt}.

\textbf{Prompt Learning for Cross-Domain Tasks.}
The prompt learning method, through the adjustment and optimization of prompts, can adapt to the language and characteristics of different domains. Consequently, it has shown promising results in cross-domain recommendation tasks \cite{ge2023domain, liu2022prompt, wu2022adversarial, zhao2023spc, zhao2022adpl}. In these tasks, shared knowledge from source domains is often transferred to the target domain via knowledge-enhanced prompts. Unlike hard prompts, which require extensive handcrafting and are highly specific to individual tasks, soft prompts offer greater flexibility and can be more easily optimized and adapted \cite{zhao2023spc}. This adaptability reduces inefficiencies and improves robustness across various tasks and models. Hard prompts, on the other hand, often present significant challenges; poorly designed prompts can negatively impact model performance and may not transfer effectively across tasks \cite{zhou2022large}. Moreover, the need for prompt engineering and the difficulty of creating effective templates for each task further limit their efficiency \cite{liu2022design, sanh2021multitask}. Therefore, this paper focuses on leveraging soft prompt learning for the CDCD task.

\section{Conclusion}
In this paper, we proposed the~\shortname~framework for cross-domain cognitive diagnosis tasks in intelligence education. Specifically, we designed the prompts to enhance the student and exercise representation across domains. The prompts follow a two-stage mode of pre-training and fine-tuning. Importantly, the proposed framework can be applied to both student-aspect and exercise-aspect cross-domain scenarios. Experimental results on the real-world datasets illustrated the effectiveness of the proposed framework.
\newpage
\bibliographystyle{unsrt}
\bibliography{acmart}

\newpage

\section*{Supplements}

\subsection*{Pseudocodes of PromptCD-S and PromptCD-E}
We present the pseudocodes of PromptCD-S and PromptCD-E applied to NeuralCD, as shown in Algorithms \ref{alg_promptCDS} and \ref{alg_promptCDE}.

\begin{algorithm}[h]
\small
  \renewcommand{\algorithmicrequire}{\textbf{Input:}}
\renewcommand\algorithmicensure {\textbf{Output:}}
  \caption{PromptCD-S for NeuralCD}
  \label{alg_promptCDS}
  \begin{algorithmic}[1]
\STATE \textbf{Input:}  The cognitive diagnosis model $\mathcal{M}$ based on the NeuralCD backbone. $\mathcal{M}$, records $\boldsymbol{LS}$ for pre-training and $\boldsymbol{LT}_{t}^{few}$ in target doamin $t$ for fine-tuning.
    \STATE \textbf{Output:} fine-tuned model $\mathcal{M}$, the transfer prompts $\boldsymbol{\hat p}^{o}_{exer}$ and $\boldsymbol{\hat p}^{d}_{sch}$.
    \STATE ---\textbf{Pre-training Stage}---
    \WHILE{$e_{1} \leqslant Epoch_{Pretrain}$}
    \FOR{$\boldsymbol{LS}_{s} \in \{\boldsymbol{LS}_{1}, \boldsymbol{LS}_{2},...,\boldsymbol{LS}_{|\boldsymbol{S}|}$\}}
    \STATE Initialize the students embedding $\boldsymbol{\alpha}_{s}^{\text{orig}}$, the exercises embedding $\boldsymbol{\beta}_{s}^{\text{orig}}$,  prompts $\boldsymbol{p}^{o}_{exer}$, $\boldsymbol{p}^{d}_{sch}$ and $\mathcal{M}$;
    \STATE Enhance the representation of students and exercises in Eq.\eqref{eq_source_con} and Eq.\eqref{eq_source_fc}, connect $\boldsymbol{p}^{o}_{exer}$ to $\boldsymbol{\beta}_{s}^{\text{orig}}$, and $\boldsymbol{p}^{d}_{sch}$ to $\boldsymbol{\alpha}_{s}^{\text{orig}}$, obtaining $\boldsymbol{\alpha}_{s}^{\text{out}}$ and $\boldsymbol{\beta}_{s}^{\text{out}}$ after mapping;
    \STATE Input $\boldsymbol{\alpha}_{t}^{\text{out}}$ and $\boldsymbol{\beta}_{t}^{\text{out}}$ to $\mathcal{M}$. Specifically, subtract $\boldsymbol{\beta}_{t}^{\text{out}}$ from $\boldsymbol{\alpha}_{t}^{\text{out}}$, multiply by the exercise and knowledge vectors, and pass through NeuralCD to get the predicted score $\boldsymbol{y}_{s}$;
    \STATE Calculate the loss using $\boldsymbol{LS}_{s}$ to update the model;
    \ENDFOR
    \ENDWHILE
    \STATE ---\textbf{Fine-Tuning Stage}---
    \WHILE{$e_{2} \leqslant Epoch_{Finetune}$}
    \STATE Initialize the students embedding $\boldsymbol{\alpha}_{t}^{\text{orig}}$, the exercises embedding $\boldsymbol{\beta}_{t}^{\text{orig}}$;
    \STATE Obtain the transfer prompts $\boldsymbol{\hat p}^{o}_{exer}$ and $\boldsymbol{\hat p}^{d}_{sch}$ in Eq.\eqref{eq_target_O_transfer} and Eq.\eqref{eq_target_D_transfer});
    \STATE Activate improvement policy in Eq.\eqref{eq_target_O_transform};
    \STATE Enhance the representations in a manner similar to pre-training;
    \STATE Input $\boldsymbol{\alpha}_{t}^{\text{out}}$ and $\boldsymbol{\beta}_{t}^{\text{out}}$ to $\mathcal{M}$ to obtain the final predicted score $\boldsymbol{y}_{t}$, as in the pre-training process;
    \STATE Calculate the loss using $\boldsymbol{LT}_{t}^{few}$ to update the model.
  \ENDWHILE
  \end{algorithmic}
\end{algorithm}

\begin{algorithm}[h]
\small
  \renewcommand{\algorithmicrequire}{\textbf{Input:}}
\renewcommand\algorithmicensure {\textbf{Output:}}
  \caption{PromptCD-E for NeuralCD}
  \label{alg_promptCDE}
  \begin{algorithmic}[1]
\STATE \textbf{Input:}  The cognitive diagnosis model $\mathcal{M}$ based on the NeuralCD backbone. $\mathcal{M}$, records $\boldsymbol{LS}$ for pre-training and $\boldsymbol{LT}_{t}^{few}$ in target doamin $t$ for fine-tuning.
    \STATE \textbf{Output:} fine-tuned model $\mathcal{M}$, the transfer prompts $\boldsymbol{\hat p}^{o}_{stu}$ and $\boldsymbol{\hat p}^{d}_{sub}$.
    \STATE ---\textbf{Pre-training Stage}---
    \WHILE{$e_{1} \leqslant Epoch_{Pretrain}$}
    \FOR{$\boldsymbol{LS}_{s} \in \{\boldsymbol{LS}_{1}, \boldsymbol{LS}_{2},...,\boldsymbol{LS}_{|\boldsymbol{S}|}$\}}
    \STATE Initialize the students embedding $\boldsymbol{\alpha}_{s}^{\text{orig}}$, the exercises embedding $\boldsymbol{\beta}_{s}^{\text{orig}}$,  prompts $\boldsymbol{p}^{o}_{stu}$, $\boldsymbol{p}^{d}_{sub}$ and $\mathcal{M}$;
    \STATE Enhance the representation of student and exercise in Eq.\eqref{eq_source_con} and Eq.\eqref{eq_source_fc}; connect $\boldsymbol{p}^{o}_{stu}$ to $\boldsymbol{\alpha}_{s}^{\text{orig}}$, and $\boldsymbol{p}^{d}_{sub}$ to $\boldsymbol{\beta}_{s}^{\text{orig}}$, obtaining $\boldsymbol{\alpha}_{s}^{\text{out}}$ and $\boldsymbol{\beta}_{s}^{\text{out}}$ after mapping.
    \STATE Input $\boldsymbol{\alpha}_{s}^{\text{out}}$ and $\boldsymbol{\beta}_{s}^{\text{out}}$ into $\mathcal{M}$. Specifically, subtract $\boldsymbol{\beta}_{t}^{\text{out}}$ from $\boldsymbol{\alpha}_{t}^{\text{out}}$, multiply by the exercise and knowledge vectors, and pass through NeuralCD to get the predicted score $\boldsymbol{y}_{s}$;
    \STATE Calculate the loss using $\boldsymbol{LS}_{s}$ to update the model;
    \ENDFOR
    \ENDWHILE
    \STATE ---\textbf{Fine-Tuning Stage}---
    \WHILE{$e_{2} \leqslant Epoch_{Finetune}$}
    \STATE Initialize the students embedding $\boldsymbol{\alpha}_{t}^{\text{orig}}$, the exercises embedding $\boldsymbol{\beta}_{t}^{\text{orig}}$;
    \STATE Obtain the transfer prompts $\boldsymbol{\hat p}^{o}_{stu}$ and $\boldsymbol{\hat p}^{d}_{sub}$ in Eq.\eqref{eq_target_O_transfer} and Eq.\eqref{eq_target_D_transfer});
    \STATE Activate improvement policy in Eq.\eqref{eq_target_O_transform};
    \STATE Enhance the representations in a manner similar to pre-training;
    \STATE Input $\boldsymbol{\alpha}_{t}^{\text{out}}$ and $\boldsymbol{\beta}_{t}^{\text{out}}$ to $\mathcal{M}$ to obtain the final predicted score $\boldsymbol{y}_{t}$, as in the pre-training process;
    \STATE Calculate the loss using $\boldsymbol{LT}_{t}^{few}$ to update the model.
  \ENDWHILE
  \end{algorithmic}
\end{algorithm}

\newpage

\vfill

\end{document}